\documentclass[lettersize,journal]{IEEEtran}
\usepackage{amsmath,amsfonts}
\usepackage{algorithm}
\usepackage[noend]{algorithmic}

\usepackage{array}
\usepackage[caption=false,font=normalsize,labelfont=rm,textfont=rm]{subfig}
\usepackage{textcomp}
\usepackage{hyperref}
\usepackage{booktabs}
\usepackage{pifont}
\usepackage{amssymb}
\usepackage{stfloats}
\usepackage{url}
\usepackage{verbatim}
\usepackage{graphicx}
\usepackage{cite}
\usepackage{color}
\usepackage[justification=justified]{caption} %

\usepackage{booktabs} %
\usepackage{graphicx} %
\usepackage{lipsum}   %
\newcommand{\XSolidBrush}{\ding{55}}      %
\newcommand{\CheckmarkBold}{\ding{52}}    %
\usepackage{multirow} 
\usepackage{makecell}

 \usepackage{xcolor}
\definecolor{c1}{HTML}{1F9856}
\definecolor{c2}{HTML}{BE0D16}
\usepackage{bm}
\usepackage[normalem]{ulem}
\usepackage{cases}
\usepackage{multirow}
\usepackage{algorithm}
\usepackage{xr}
\usepackage[noend]{algorithmic}
\renewcommand{\algorithmicrequire}{}  
\usepackage{nicematrix} %

\usepackage{xcolor}

\renewcommand{\algorithmicrequire}{}   

\definecolor{my1}{RGB}{0,112,204}
\definecolor{my2}{RGB}{0,128,121}

\begin{document}
\title{Towards Efficient and Exact Forgetting Services in Pre-Trained-Model-based Continual Learning}

\author{
Yajiang Huang, 
Jianheng Tang,
Kejia Fan,
Huiping Zhuang,
Anfeng Liu,Tian Wang,~\IEEEmembership{Senior Member,~IEEE},\\
Yunhuai Liu,~\IEEEmembership{Member,~IEEE},
Mianxiong Dong,~\IEEEmembership{Senior Member,~IEEE},
Houbing Song,~\IEEEmembership{Fellow,~IEEE}

\thanks{Manuscript received XXXX XXXXXX, XXXX; revised XXXX XXXXXX, XXXX; accepted XXXX XXXXXX, XXXX. 
% This work is supported in part by the National Key Research and Development Program of China under No. 2024YFC2607404, the National Natural Science Foundation of China under Nos. U24A20248 and 62576013.
We also sincerely thank  Di Fang, Jiaxu Li, Feijiang Han, Leye Wang, Zhanxing Zhu, and Shanghang Zhang for their valuable contributions to this paper.
% (Corresponding author: Anfeng Liu).
}

\thanks{Yajiang Huang, Kejia Fan, and Anfeng Liu are with the School of Computer Science and Engineering, Central South University, Changsha 410083, China. (e-mail: yjhuang@csu.edu.cn,
kejiafan@csu.edu.cn, 
afengliu@mail.csu.edu.cn).}%
\thanks{Jianheng Tang and Yunhuai Liu are with the School of Computer Science, Peking University, Beijing, 100871, China, and are also with the Key Lab of High Confidence Software Technologies (Peking University), Ministry of Education, China. (e-mail: tangentheng@gmail.com, yunhuai.liu@pku.edu.cn).}
\thanks{Huiping Zhuang is with the Shien-Ming Wu School of Intelligent Engineering, South China University of Technology, Guangzhou 510641, China. (e-mail: hpzhuang@scut.edu.cn).}
\thanks{Tian Wang is with the Department of Artificial Intelligence and Future Networks, Beijing Normal University, Zhuhai, Guangdong, 519087, China. (email: tianwang@bnu.edu.cn).}
\thanks{Mianxiong Dong is with the Department of Information and Electronic Engineering, Muroran Institute of Technology, Muroran, 050-8585, Japan. (e-mail: mx.dong@csse.muroran-it.ac.jp).}
\thanks{Houbing Herbert Song is with the Department of Information Systems, University of Maryland, Baltimore County (UMBC), Baltimore, MD 21250 USA. (email: h.song@ieee.org).}
}

\markboth{}%
{How to Use the IEEEtran \LaTeX \ Templates}

\maketitle

\begin{abstract}

In Continual Learning (CL), using a Pre-Trained Model (PTM) as the feature extractor has become a popular practice.
Accompanied by analytic classifiers, the PTM-based methods have achieved state-of-the-art performance in CL, in pursuit of the non-forgetting goal.
Meanwhile, actively forgetting specific knowledge acquired during the CL phase is also essential in most service construction paradigms, for example, Mobile Crowd Sensing (MCS), where mobile edge nodes continuously collect sensory data and demand not only non-forgetting adaptation but also specific knowledge forgetting for privacy preservation.
Thus, a unique problem, called Continual Unlearning (CU), arises when the forgetting requests show sequentially in CL.
However, existing unlearning methods focus on single-shot joint forgetting and prove highly inadequate when applied to CU, including (1) violating the historical data privacy in CL and (2) vulnerably being overwhelmed or degraded with adversarially frequent requests.
To handle the challenges of CU, we propose a gradient-free approach, called \underline{\textbf{A}}nalytic \underline{\textbf{C}}ontinual \underline{\textbf{U}}nlearning (ACU), for efficient and exact forgetting with historical data privacy preservation in PTM-based CL.
In response to each unlearning request, our ACU recursively derives the analytical (i.e., closed-form) solutions via least squares in an interpretable manner.
By meticulous design, our ACU is compatible with both sample-level and class-level unlearning requests.
The theoretical and experimental evaluations validate our ACU's superiority in unlearning effectiveness, model fidelity, and system efficiency.

\end{abstract}


\section{Introduction}
\label{sec:intro}

Continual Learning (CL) is a significant topic for machine intelligence to adapt to the open-world service construction~\cite{continual-learning-unlearning-4, TKDE-CL1, TSC-CL1}.
In particular, CL has emerged as an important service construction paradigm in many real-world applications, for example, Mobile Crowd Sensing (MCS), where mobile edge nodes continuously collect sensory data and demand continual model adaptation~\cite{CALM}.
In CL, using a Pre-Trained Model (PTM) as the feature extractor has become a popular practice~\cite{cil-ranpac, ACIL_0, CIL-survey, CIL-pretrain-tool, ACIL_2}.
Paired with analytic classifiers, many PTM-based methods have achieved state-of-the-art performance in CL, in pursuit of the \textit{non-forgetting} goal~\cite{MiN, Anacp_CL, ACIL_4}.
Nevertheless, the ability to actively forget specific knowledge acquired during the CL phase (e.g., the ``\textit{right to be forgotten}”) is equally crucial in most high-quality service construction paradigms, such as MCS and federated learning for meeting privacy and security requirements  ~\cite{TKDE-Unlearning1, unlearning-1, TKDE-Unlearning2, TKDE-MCS1, TSC-MCS1, TKDE-MCS2}.

\begin{figure*}[t]
    \centering
    \includegraphics[width=1.0\linewidth]{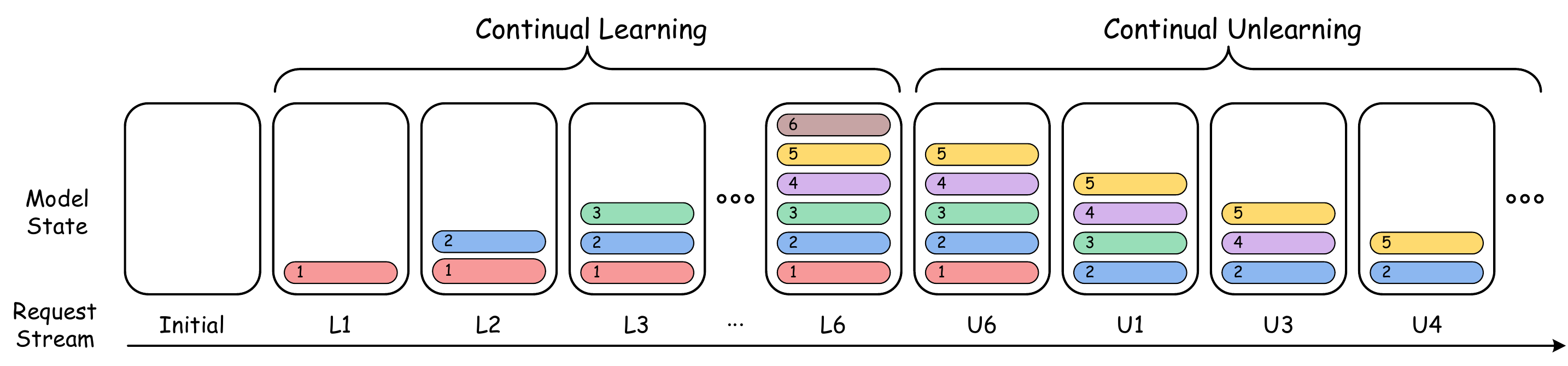}
    \caption{Illustration of model state evolution under the CL and CU phases. The model begins blank with no knowledge.  As the CL phase progresses, the model incrementally acquires knowledge from online tasks. After the CL phase, the CU requests are sequentially issued to remove particular knowledge acquired during the CL phase. Since the online data from the CL phase is discarded after training, only the designated samples to be forgotten can be accessible for each CU request.
    The CL and CU phases can iterate in an alternating manner.
    }
    \label{fig:fig-1-continual-unlearning}
    \vspace{-0.15cm}
\end{figure*}

In this context, a unique and challenging problem, which we term Continual Unlearning (CU), arises when forgetting requests are issued sequentially in CL, as shown in Figure~\ref{fig:fig-1-continual-unlearning}.
CU has two major characteristics:
(1) Its forgetting target is the knowledge acquired in CL, requiring adherence to CL's privacy constraints, which state that the historical data should not be accessible;
(2) It can involve multiple and different forgetting requests that are triggered frequently.
While several recent advances have shown progress in machine unlearning, they mainly focus on single-shot joint forgetting and are highly inadequate when applied to our introduced CU problem.

First of all, the existing unlearning methods typically assume that the original model is trained in a centralized and joint manner, with access to the retained data for \textit{re-training} or \textit{fine-tuning}~\cite{TKDE-Unlearning1,unlearning-2-exact,TKDE-Unlearning2, TKDE-Unlearning3}.
Yet, this assumption no longer holds true in CU, where the model is trained in a CL fashion, and the historical training data cannot be revisited.
Hence, the only available information for each unlearning request is limited to the designated samples/classes to be forgotten.

Meanwhile, even if the historical data are fully available and privacy concerns are ignored, frequent triggering of requests during CU still poses a key challenge.
Specifically, the {\textit{cumulative effect}} of multiple unlearning requests in CU can severely impair the \textit{model fidelity} and \textit{system efficiency}.
As a result, existing unlearning methods often exhibit a poor trade-off between these two indicators, especially under the stress of adversarially frequent triggering~\cite{TSC-unlearing2,unlearning-8,unlearning-2-exact}. 

On the one hand, exact unlearning methods typically rely on partial re-training via the retained dataset, which incurs substantial computational and time overhead~\cite{unlearning-2-exact,unlearning-4-exact,unlearning-5-exact}.
The accumulated cost in CU creates a harsh vulnerability to the Denial-of-Service (DoS) attacks, as a single forgetting target may be fragmented into multiple frequent requests to overwhelm the system.
On the other hand, the methods with more efficiency are limited to approximate unlearning, risking unintended removal of valuable knowledge that ought to be preserved and compromising the model fidelity on non-forgotten data~\cite{unlearning-9,TKDE-Unlearning2,unlearning-11}.
Such compromises in model fidelity can accumulate and compound during CU, leading to catastrophic degradation or even collapse of the model, reducing the quality of service.

In summary, existing methods have two key limitations in CU: (1) the prevailing reliance on the historically retained data, and (2) the poor trade-off between system efficiency and model fidelity.
In this paper, we identify that these limitations of existing methods fundamentally stem from their reliance on gradient-based model updates.
Specifically, the gradients render learning from data a complex, entangled, and incremental process with inherent stochasticity, making it hard to explicitly characterize the influence of data on the model, let alone exactly erase it.
It also forms the reason why existing methods require access to the retained data for recovering model fidelity.
Several studies have recognized the flaws of gradient-based updates, but few have attempted to fundamentally address this issue at its root~\cite{TSC-unlearing1, TSC-unlearing2, unlearning-hard-1, unlearning-hard-2}.

Currently, the community of CL has witnessed the timely emergence of gradient-free techniques in many PTM-based CL methods, particularly grounded in the use of the analytic classifiers~\cite{ACIL_0, ACIL_3, ACIL_1, ACIL_2, MiN, Anacp_CL, CALM, CFSSeg, GraphKeeper, ACIL_L3A, new1, Any-SSR,ACIL_4}.
These PTM-based methods with analytic classifiers have indicated the ideal \textit{non-forgetting} (also called \textit{absolute memorization}) property due to the avoidance of gradient-based updates.
On this basis, we aim to explore the gradient-free power of the analytic classifiers in CU, further showing its \textit{free-forgetting} property, i.e., retaining all knowledge that should not be forgotten and erasing all knowledge designated for unlearning.

To achieve this goal and bridge the gaps in CU, in this paper, we propose a gradient-free approach, called \underline{\textbf{A}}nalytic \underline{\textbf{C}}ontinual \underline{\textbf{U}}nlearning (ACU), for efficient and exact forgetting with historical data privacy preservation in PTM-based CL.
Our ACU can be directly plugged into the existing CL methods with analytic classifiers, enabling accurate and efficient erasure of designated knowledge acquired during the CL phase.
In response to each unlearning request, our ACU recursively derives analytical (i.e., closed-form) solutions via least squares in an interpretable manner.
By meticulous design, our ACU is compatible with both sample-level and class-level unlearning requests.
Hence, our ACU can boost better trustworthiness and responsibility in PTM-based CL.

Our main contributions are summarized as follows:
\begin{enumerate}
    \item Conceptually, we introduce a new problem, namely CU, analyze its key challenges, and identify the fundamental issue of existing methods as their reliance on gradients.
    \item Technically, we devise an efficient gradient-free protocol for our ACU, providing a plug-and-play solution for unlearning in existing Analytic Continual Learning (ACL) methods, without replaying any historical data.
    \item Theoretically, we validate our ACU's exact unlearning capabilities, establishing that it yields identical results to those from joint re-training on the remaining datasets.
    Moreover, we also show our ACU's excellent efficiency.
    \item Experimentally, we confirm the superiority of our ACU in terms of unlearning effectiveness, model fidelity, and system efficiency.
    In particular, it exhibits progressively pronounced advantages as the CU process advances.
\end{enumerate}

\section{Related Works}
\label{sec:related}

\subsection{Machine Unlearning}

Recently, machine unlearning has emerged as a popular research topic, aiming to delete specific data and eliminate the corresponding influence on the learned models~\cite{TSC-unlearing1, unlearning-3, unlearning-certified}.
This topic was originally motivated by the need to meet privacy and regulatory requirements (e.g., upholding the users'  ``\textit{right to be forgotten}”)~\cite{TSC-unlearing2, unlearning-1, unlearning-3, TKDE-Unlearning2}.
Later, its scope has been expanded to include the active removal of erroneous or biased knowledge derived from toxic data, thereby ensuring the trustworthiness of learned models~\cite{unlearning-1,unlearning-LLM,Continual-forgetting,unlearning-survey-federated}.
Existing unlearning methods typically assume that the original model is trained in a centralized joint manner, with access to the retained data for re-training or fine-tuning~\cite{TSC-unlearing1,unlearning-2-exact,TKDE-Unlearning2}.
Yet, such an assumption no longer holds valid in CU, where the model is trained in a CL fashion and the historical training data cannot be revisited.
Moreover, existing unlearning methods mainly focus on single-shot joint forgetting, thus suffering from the poor trade-off between system efficiency and model fidelity under frequent unlearning requests~\cite{unlearning-continual-AAAI,continual-learning-unlearning-1}.

Although there have been a few works attempting to connect CL with unlearning~\cite{continual-learning-unlearning-survey,continual-learning-unlearning-1,continual-learning-unlearning-2,continual-learning-unlearning-3,continual-learning-unlearning-4,continual-learning-unlearning-5}, their focus differs significantly from our introduced CU problem.
Among them, ErrorEraser~\cite{continual-learning-unlearning-4} uses intentional forgetting to enhance CL performance, but does not handle external unlearning requests.
UniCLUN~\cite{continual-learning-unlearning-1} achieves both CL and unlearning within a unified framework via knowledge distillation. 
Yet, UniCLUN~\cite{continual-learning-unlearning-1} requires replaying historical data and cannot achieve exact forgetting.
CLPU~\cite{continual-learning-unlearning-2} learns independent networks for each task and discards them on request.
While CLPU~\cite{continual-learning-unlearning-2} achieves exact unlearning, this naive approach comes at the expense of a sharp increase in the number of parameters.
UnCLe~\cite{continual-learning-unlearning-3} uses hypernetworks for CL and approximate unlearning without historical data.
Nevertheless, both CLPU~\cite{continual-learning-unlearning-2} and UnCLe~\cite{continual-learning-unlearning-3} are limited to task-level forgetting granularity (i.e., all samples of an entire task in CL), failing to handle fine-grained sample/class-level unlearning that partially arises within one CL task.
GS-LoRA~\cite{Continual-forgetting,Continual-forgetting2} proposes continual forgetting for the pre-trained backbones without involving CL, thus standing orthogonal to our ACU.
In summary, CU is a new problem distinct from all previous studies, and our ACU is the first work to fulfill efficient and exact forgetting with historical data privacy preservation.

\begin{table}[t]
    \centering
    \small
    \renewcommand{\arraystretch}{1.15}
    \caption{Description of important notations}
    \label{table-notation}
    \begin{NiceTabular}{
        >{\centering\arraybackslash}p{0.15\columnwidth}
        >{\raggedright\arraybackslash}p{0.75\columnwidth}
    }
        \toprule
        \textbf{Notations} & \textbf{Description} \\
        \midrule
        $\mathcal{D}$ 
        & The complete training dataset during the CL phase. \\

        $\mathcal{\check D}_i$ 
        & The forgetting dataset corresponding to the $i$-th unlearning request. \\

        $\mathcal{\hat D}_i$ 
        & The retained dataset after removing the first $i$ forgetting datasets from $\mathcal{D}$. \\

        $\mathbf{f}_j$ 
        & The feature vector for the $j$-th data sample. \\

        $\mathbf{y}_j$ 
        & The label vector for the $j$-th data sample. \\

        $\mathbf{W}_0$ 
        & The initial analytic model of the CU phase, i.e., the final model of the CL phase. \\

        $\mathbf{W}_i$ 
        & The updated analytic model after processing the $i$-th unlearning request. \\

        $\mathbf{\hat W}_i$ 
        & The re-trained model on the retained dataset $\mathcal{\hat D}_i$. \\

        $\mathbf{T}_i$ 
        & The \textit{Knowledge Tracking Matrix} after processing the $i$-th unlearning request. \\
        \bottomrule
    \end{NiceTabular}
\end{table}

\subsection{Analytic Learning}

Analytic learning (also termed pseudoinverse learning) is a representative gradient-free technique, originally developed to address the common gradient-related issues, e.g., vanishing and exploding gradients~\cite{ACIL_1, AL_1, AL_new_0, AL_new_1,AL_3,AL_5}.
Its core idea is to directly derive the analytical (i.e., closed-form) solutions via least squares~\cite{ACIL_2, CALM, CFSSeg}. 
With the adoption of the block-wise recursive Moore-Penrose inverse~\cite{AL_6}, analytic learning has shown vast superiority in many applications, particularly achieving state-of-the-art performance in PTM-based CL~\cite{new1, AFL, ACIL_4, CFSSeg, GraphKeeper}.
Those PTM-based CL methods using analytic classifiers are collectively unified under the term of Analytic Continual Learning (ACL)~\cite{ACIL_0, ACIL_1, ACIL_2, ACIL_3}.

Currently, studies on ACL primarily focus on leveraging its ideal \textit{non-forgetting} (also called \textit{absolute memorization}) property to further enhance its upper-bound performance or broaden its application scope in CL~\cite{ACIL_0, ACIL_3, ACIL_1, ACIL_2, ACIL_4, MiN, Anacp_CL, CALM, CFSSeg, GraphKeeper, ACIL_L3A, new1, Any-SSR}.
Yet, to the best of our knowledge, there is no existing work exploring the potential of analytic classifiers in unlearning.
Therefore, by introducing CU as a new, significant, and challenging problem, we aim to explore the inverse counterpart of ACL beyond resisting forgetting, i.e., facilitating exact unlearning.
Specifically, we show that the analytic classifiers can not only preserve all knowledge that should not be forgotten, but also erase all knowledge that is designated for unlearning, thereby extending their property from prior \textit{non-forgetting} to \textit{free-forgetting}.
In conjunction with the broad applications of ACL, our work can further enhance its trustworthiness and responsibility in practice.

\section{Our Proposed ACU Method}
\label{sec:method}

\subsection{System Model}
As shown in Figure~\ref{fig:fig-1-continual-unlearning}, the CU phase aims to continuously respond to unlearning requests and forget specific knowledge acquired during the CL phase.
Specifically, we denote the original model at the beginning of the CU phase as $\mathbf{W}_0$, which corresponds to the final model obtained from the CL phase.
We use $\mathcal{D} = {\{(x_j, y_j)\}}_{j=1}^N$ to denote the full training dataset of the CL phase for $\mathbf{W}_0$, where $x_j$ and $y_j$ denote the input and label of the $j$-th sample. 
The data required to be forgotten in the unlearning requests are denoted as a sequence of forgetting sets $\mathcal{\check D} = \{\mathcal{\check  D}_1, \cdots, \mathcal{\check D}_i, \cdots\}$, where we have $\mathcal{\check D}_i \subseteq \mathcal{D}$ and $\mathcal{\check D}_i \cap \mathcal{\check D}_{i'} = \emptyset$ for $\forall i \neq i'$.

We denote the unlearned model after processing the first $i$ unlearning requests as $\mathbf{W}_i$.
The goal for each unlearning request $i$ is to update the current model $\mathbf{W}_{i-1}$ for obtaining $\mathbf{W}_i$, such that $\mathbf{W}_i$ is as close as possible to the model re-trained from scratch on the retained set $\mathcal{\hat D}_i = \mathcal{D} \setminus \bigcup_{k=1}^i \mathcal{\check D}_k$ (excluding all data to be forgotten), as if the unlearned data had never been used in training.
Notably, the retained set $\mathcal{\hat D}_i$ is inaccessible in practice, because the online task data are all discarded once training is completed during the CL phase. 
Thus, the only available information is limited to the designated samples to 
be forgotten, i.e., $\mathcal{\check  D}_i$.
For clarity, we summarize our primary notations in Table \ref{table-notation}.

\subsection{Framework Overview}

\begin{figure*}[t]
    \centering
    \includegraphics[width=1.0\linewidth]{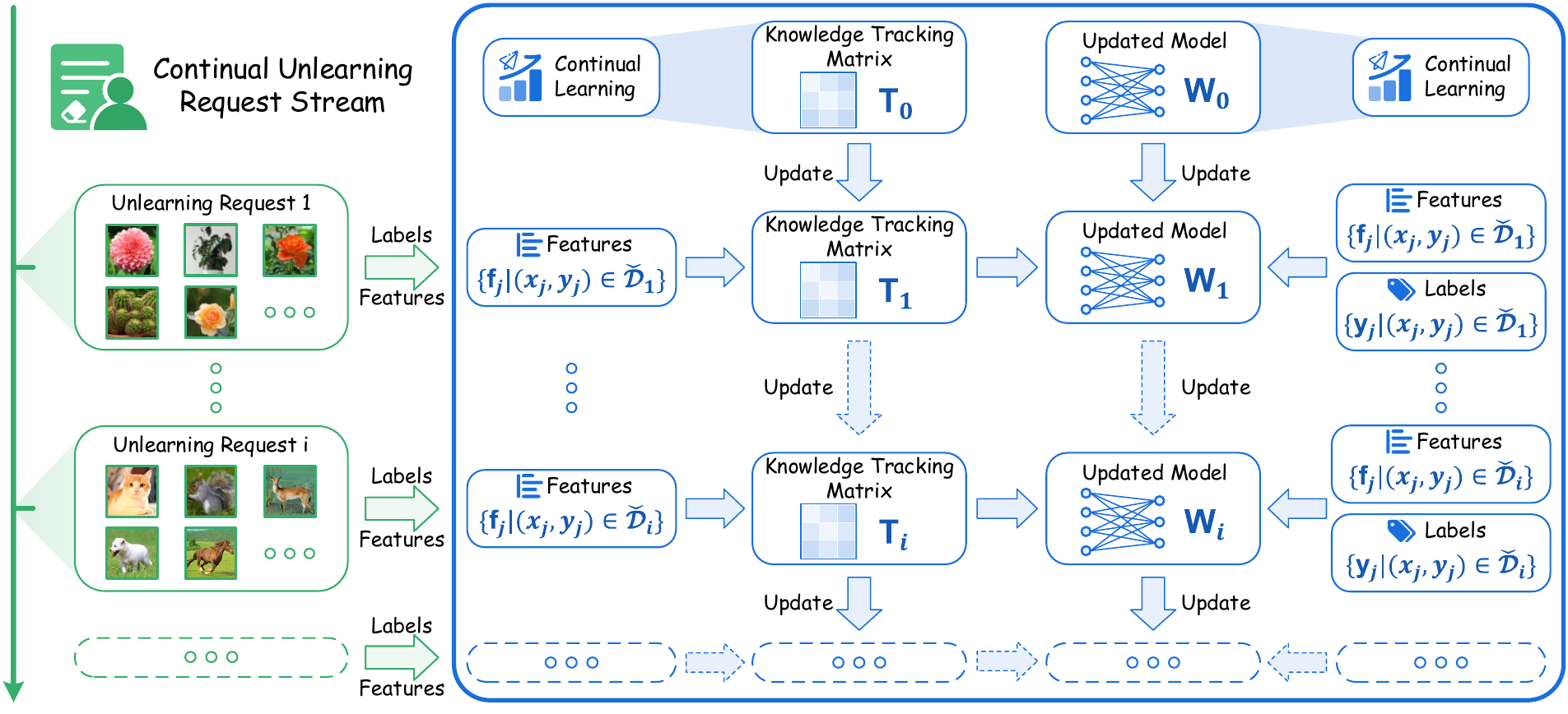}
    \caption{The framework overview of our proposed ACU.}
    \label{fig:detail-framwork}
    \vspace{-0.15cm}
\end{figure*}

In this paper, we employ image classification as a representative example to illustrate ACU's workflow.
Notably, ACL methods have been successfully applied to various domains.
As a plug-and-play solution for unlearning in existing ACL methods, ACU can be readily extended to these domains in future work, such as graph learning~\cite{GraphKeeper}, semantic segmentation~\cite{CFSSeg}, and multimodal learning~\cite{ACIL_5}.
Aligned with existing ACL methods~\cite{exemplar-free-0, exemplar-free-1, ACIL_0, ACIL_3}, the model within ACU consists of two key components: a PTM-based feature extractor and a gradient-free analytic classifier.
Specifically, the PTM-based feature extractor $\mathrm{Backbone}( \cdot , \Theta)$ and the one-hot function $\mathrm{Onehot}( \cdot)$ are employed to derive the feature and label vectors for each sample in $\mathcal{D}$, as follows:
\begin{equation}
\label{eq:backbone}
    \mathbf{f}_j = \mathcal G(\mathrm{Backbone}(x_j, \Theta)), \quad
    \mathbf{y}_j = \mathrm{Onehot}(y_j).
\end{equation}

Here, $\mathbf{f}_j \in \mathbb{R}^{1 \times d_\mathrm{F}}$ and $\mathbf{y}_j \in \mathbb{R}^{1 \times d_\mathrm{C}}$ denotes the obtained feature vector and label vector, where $d_\mathrm{F}$ and $d_\mathrm{C}$ represents the feature dimension and the total number of classes.
Notably, as is already stated in many PTM-based CL methods, the PTM $\mathrm{Backbone}( \cdot , \Theta)$ can be easily sourced from public models on the Internet~\cite{CIL-survey, exemplar-free-0, ACIL_new-1, ACIL_L3A, CIL-pretrain-tool}.
Furthermore, the nonlinear mapping $\mathcal{G}(\cdot)$ in \eqref{eq:backbone} can be implemented using \textit{random projections} or \textit{kernel functions} to enhance the representational capacity (e.g., linear separability) of features, as a widely adopted practice in the CL community~\cite{ACIL_0, ACIL_3, ACIL_2, ACIL_1}.

Then, the analytic classifier $\mathbf{W}_{i}$ is appended to the PTM-based feature extractor to generate the classification results.
Specifically, for any analytic classifier $\mathbf{W}_{i}$ in the CU phase, the optimization objective is formulated by minimizing the Mean Squared Error (MSE) with regularization over the retained set $\mathcal{\hat D}_i$, as shown in \eqref{eq:Motivation-1}.
This formulation aligns with the training objective of the ACL methods~\cite{ACIL_L3A,MiN,ACIL_4}.
\begin{equation}
\label{eq:Motivation-1}
    \mathbf{W}_{i}  = 
\mathop{\arg\min}\limits_{\mathbf{W}}\sum\nolimits_{j:(x_j,y_j) \in \mathcal{\hat D}_i} {\left\|\mathbf{y}_j -\mathbf{f}_j \mathbf{W} \right\|}_\mathrm{F}^2 + \gamma{\left\| \mathbf{W} \right\|}_\mathrm{F}^2. 
\end{equation}

As a special case, when $i=0$, $\mathcal{\hat D}_0 = \mathcal{D}$ and $\mathbf{W}_0$ represents both the initial model of the CU phase and the final model of the CL phase.
Despite the classification task, we deliberately adopt MSE loss over the commonly used Cross-Entropy (CE) loss in \eqref{eq:Motivation-1} to enable a closed-form solution, which is essential for gradient-free model updates.
Meanwhile, the primary advantage of the CE lies in its ability to facilitate convergence in the context of gradient-based updates, which no longer holds in the context of gradient-free updates.
In fact, recent work has broadly shown that MSE can yield comparable performance to CE~\cite{MSE-0, MSE-1}, with better robustness to combat catastrophic forgetting in CL~\cite{MiN, ACIL_2, ACIL_new-1}.

As shown in Figure~\ref{fig:detail-framwork}, to enable our ACU to continually unlearn the specific knowledge, we design the \textit{Knowledge Tracking Matrices} to assist the CU phase.
Specifically, upon receiving each unlearning request $i$, our ACU first extracts the labels and features of the corresponding forgetting set $\mathcal{\check D}_i$ using \eqref{eq:backbone}.
Subsequently, we can recursively update the \textit{Knowledge Tracking Matrix} $\mathbf{T}_i$ and the analytic classifier $\mathbf{W}_i$ to achieve efficient and exact forgetting with historical data privacy preservation in a gradient-free manner.

\subsection{Analytic Knowledge Embedding}
\label{subsection-3.2-Traceable Knowledge Embedding}

In this subsection, we concentrate on exploring the analytic knowledge embedding during the CL phase, which is the foundation for performing exact forgetting in the subsequent CU phase. 
Specifically, for the objective \eqref{eq:Motivation-1}, we can derive a closed-form solution of $\mathbf{W}_0 \in \mathbb{R}^{d_\mathrm{F} \times d_\mathrm{C}}$ via least squares, as established in \eqref{eq:subsection-3.2-2}.
This formulation explicitly embeds all the knowledge acquired during the CL phase and serves as the model initialization for the CU phase. 
For focus and clarity, we defer the theoretical derivation for transforming \eqref{eq:Motivation-1} into the formula \eqref{eq:subsection-3.2-2} to Section~\ref{subsection-3.4-Theoretical Analyses}.
\begin{equation}
\label{eq:subsection-3.2-2}
    \mathbf{W}_0 = ( \sum\nolimits_{j:(x_j,y_j) \in \mathcal{D}} \mathbf{f}_j^\top \mathbf{f}_j +\gamma\mathbf{I})^{-1} \sum\nolimits_{j:(x_j,y_j) \in \mathcal{D}} \mathbf{f}_j^\top \mathbf{y}_j.
\end{equation}

Here, we do not delve into the details of how to obtain the final model $\mathbf{W}_{0}$ during the CL phase that satisfies \eqref{eq:subsection-3.2-2}, as many existing ACL methods already solved this problem by using recursive formulas and achieved the state-of-the-art performance~\cite{ACIL_L3A, CFSSeg, ACIL_1, ACIL_2, ACIL_new-1, MiN}.
Instead, we aim to characterize the influence of each data sample on the model $\mathbf{W}_{0}$ based on its closed-form solution \eqref{eq:subsection-3.2-2}.
This aim is nearly impossible in traditional gradient-based methods, highlighting the gradient-free power of our ACU in unlearning.
Since we cannot access the full retained data in CU, we design a \textit{Knowledge Tracking Matrix} to track the compressed influence of all retained data samples, thereby serving as a surrogate for the private retained data.
Specifically, the initial \textit{Knowledge Tracking Matrix} $\mathbf{T}_0  \in \mathbb{R}^{d_\mathrm{F} \times d_\mathrm{F}}$ is defined as in \eqref{eq:subsection-3.2-4}, which will be updated in the CU phase.
\begin{equation}
\label{eq:subsection-3.2-4}
    \mathbf{T}_0 = (\sum\nolimits_{j:(x_j,y_j) \in \mathcal{D}} \mathbf{f}_j^\top \mathbf{f}_j + \gamma\mathbf{I})^{-1}.
\end{equation}

Notably, although \eqref{eq:subsection-3.2-4} represents the initial \textit{Knowledge Tracking Matrix} tracking all samples, it can be recursively computed during the CL phase according to the online data increments, thereby eliminating the need to centrally store all data samples $\mathcal{D}$.
More details on the feasibility of the recursive computation for $\mathbf{T}_0$ during the CL phase are provided in Appendix D.
Furthermore, since the rank of our designed \textit{Knowledge Tracking Matrix} is $R \leq d_\mathrm{F} \ll N$, it represents a highly compressed trace of the historical examples and cannot be inverted to recover the original dataset, thereby preserving historical data privacy.
More details on privacy analyses are provided in Section~\ref{subsection-3.4-Theoretical Analyses}.

\begin{algorithm}[t]
\renewcommand{\algorithmicrequire}{\textbf{Input:}}
\renewcommand{\algorithmicensure}{\textbf{Output:}}
\renewcommand{\algorithmicreturn}{\textbf{Return:}}
\begin{algorithmic}[1]
\caption{ACU for Sample-level Unlearning Requests}
\label{alg:ACU-sample}
\REQUIRE $\mathbf{W}_0$, $\mathbf{T}_0$, and forgetting sets $\mathcal{\check D} = \{\mathcal{\check  D}_1, \mathcal{\check D}_2, \cdots, \mathcal{\check D}_K \}$.
\ENSURE The final model $\mathbf{W}_K$.
\FOR {each unlearning request $i$} 
    \STATE Extract features $\mathbf{f}_j$ and labels $\mathbf{y}_j$ of the data samples within $\mathcal{\check D}_i$ using \eqref{eq:backbone}. 
    \STATE Update the \textit{Knowledge Tracking Matrix} to obtain $\mathbf{T}_i$ using \eqref{eq:subsection-3.3-1-Updating K}.
    \STATE Update the analytic classifier to obtain $\mathbf{W}_i$ using \eqref{eq:subsection-3.3-2-Unlearning}.
\ENDFOR
\RETURN The final model $\mathbf{W}_K$.
\end{algorithmic}
\end{algorithm}

\subsection{Analytic Knowledge Forgetting}
\label{subsection-3.3-Analytic Knowledge Unlearning}

In this subsection, we present the details of analytic knowledge forgetting during the CU phase.
Here, we focus on handling sample-level requests for clarity, while class-level unlearning can be regarded as forgetting all samples associated with the target class.
In practice, class-level unlearning can be achieved similarly to sample-level unlearning, yet requires additionally storing a more fine-grained \textit{Knowledge Tracking Matrix} for each class.
More details on addressing the class-level unlearning are provided in Appendix E.

Specifically, upon receiving each unlearning request $i$ along with the forgetting set $\mathcal{\check D}_i$, we first extract the feature and label vectors for all data samples within $\mathcal{\check D}_i$ using \eqref{eq:backbone}.
Based on these vectors, we sequentially update the \textit{Knowledge Tracking Matrix} to obtain $\mathbf{T}_i$ as follows.
\begin{equation}
\label{eq:subsection-3.3-1-Updating K}
    \mathbf{T}_i = \mathbf{T}_{i-1} + \mathbf{T}_{i-1} \mathbf{\check F}_i^\top ( \mathbf{I} - {\mathbf{\check F}_i} \mathbf{T}_{i-1} \mathbf{\check F}_i^\top  )^{-1} {\mathbf{\check F}_i} \mathbf{T}_{i-1},
\end{equation}
where $\mathbf{\check F}_i = \begin{bmatrix} \mathbf{f}_j \end{bmatrix}_{j:(x_j,y_j) \in \mathcal{\check D}_i} \in \mathbb{R}^{|\mathcal{\check D}_i| \times d_\mathrm{F}}$ denotes the forgetting feature matrix formed by vertically stacking all feature vectors $\mathbf{f}_j$ from the forgetting set $\mathcal{\check D}_i$.
To better demonstrate the validity of the recursive formula \eqref{eq:subsection-3.3-1-Updating K}, we provide the closed-form expression of $\mathbf{T}_i$ and the detailed theoretical derivation in Lemma 3 of Appendix A.

Subsequently, leveraging the updated \textit{Knowledge Tracking Matrix} $\mathbf{T}_i$, we aim to further update the analytic classifier $\mathbf{W}_i$ for exactly erasing the specified knowledge in the forgetting set $\mathcal{\check D}_i$.
To achieve this goal, let's delve deeper into the scenario from an \textbf{\textit{Oracle}} perspective to understand what happens when we process an unlearning request.
Specifically, if we consider the optimal classifier that satisfies \eqref{eq:Motivation-1}, the unlearning request essentially corresponds to reducing the size of the retained set.
This decremental update results in two effects: (a) the influence of the remaining samples is amplified, and (b) the influence of the forgotten samples is erased.
Inspired by these effects, we recursively update the analytic classifier $\mathbf{W}_i$ using two interpretable terms to achieve exact unlearning, as follows.
\begin{equation}
\label{eq:subsection-3.3-2-Unlearning}
\mathbf{ W}_{i} =   \underbrace{(\mathbf{I} + \mathbf{T}_i  \sum\nolimits_{\mathcal{\check D}_{i}} \mathbf{f}_j^\top \mathbf{f}_j)\mathbf{W}_{i-1}}_{(a)} - \underbrace{\mathbf{T}_i \sum\nolimits_{ \mathcal{\check D}_{i}} \mathbf{f}_j^\top \mathbf{y}_j}_{(b)}.
\end{equation}

Here, the former term within  \eqref{eq:subsection-3.3-2-Unlearning} represents amplifying the influence of the retained data by multiplying the previous $\mathbf{W}_{i-1}$ by a factor greater than $\mathbf{I}$, and the factor is related to the features to be forgotten.
The latter term explicitly reflects the process of erasing the knowledge to be forgotten.
Meanwhile, the $\mathbf{T}_i$ plays a role in adjusting knowledge.
Furthermore, Theorem 1 in Section~\ref{subsection-3.4-Theoretical Analyses} demonstrates that the analytic classifier $\mathbf{W}_{i}$ updated via \eqref{eq:subsection-3.3-2-Unlearning} is equivalent to the re-trained model directly on the retained set $\mathcal{\hat D}_i$, thus verifying its capability for exact unlearning.

\begin{algorithm}[t]
\renewcommand{\algorithmicrequire}{\textbf{Input:}}
\renewcommand{\algorithmicensure}{\textbf{Output:}}
\renewcommand{\algorithmicreturn}{\textbf{Return:}}
\begin{algorithmic}[1]
\caption{ACU for Class-level Unlearning Requests}
\label{alg:ACU-class}
\REQUIRE $\mathbf{W}_0$, $\mathbf{T}_0$, $\{\mathbf{\bar T}_c, \mathbf{\bar A}_c\}_{c=1}^{d_\mathrm{C}}$, and forgetting classes $\{c_1, c_2, \cdots, c_K\}$.
\ENSURE The final model $\mathbf{W}_K$.
\FOR {each unlearning request $i$ with class $c_i$} 
    
    \STATE Update the \textit{Knowledge Tracking Matrix} to obtain $\mathbf{T}_i$ using (E.2) in Appendix E.
    
    \STATE Update the analytic classifier to obtain $\mathbf{W}_i$ using (E.3) in Appendix E
    
    \STATE Remove the column corresponding to class $c_i$ from $\mathbf{W}_i$.
    
    \STATE Remove $\mathbf{\bar T}_{c_i}$ and $\mathbf{\bar A}_{c_i}$ from the class-level matrices.
\ENDFOR
\RETURN The final model $\mathbf{W}_K$.
\end{algorithmic}
\end{algorithm}
Due to its gradient-free nature, our ACU requires only forward-propagation and lightweight matrix computation, without the need for back-propagation, significantly improving system efficiency.
In addition, since the historical knowledge is explicitly embedded in the closed-form formulas, we can maintain model fidelity without re-training or fine-tuning on the retained dataset.
Moreover, the model within our ACU retains its optimal analytical form even after undergoing the CU phase, which lays the foundation for subsequent CL phases.
In other words, our ACU can readily support alternating CL and CU phases, greatly expanding its practical application scope and potential.
For clarity, we provide the detailed procedures of our ACU for sample-level and class-level unlearning in Algorithm \ref{alg:ACU-sample} and \ref{alg:ACU-class}, respectively.

\subsection{Theoretical Analyses}
\label{subsection-3.4-Theoretical Analyses}

In this subsection, we first thoroughly analyze our ACU's the validity.
Specifically, we derive the closed-form solution to the optimization problem \eqref{eq:Motivation-1} for the re-trained models that are obtained via the retained sets $\mathcal{\hat D}_i$, as follows.

\noindent\textbf{Lemma 1:} For any optimization problem of the form:
\vspace{-0.1cm}
\begin{equation}
\vspace{-0.1cm}
\mathbf{W}_{i}  = 
\mathop{\arg\min}\limits_{\mathbf{W}}\sum\nolimits_{j:(x_j,y_j) \in \mathcal{\hat D}_i} {\left\|\mathbf{y}_j -\mathbf{f}_j \mathbf{W} \right\|}_\mathrm{F}^2 + \gamma{\left\| \mathbf{W} \right\|}_\mathrm{F}^2,
\nonumber
\end{equation}
it admits a unique closed-form solution given by:
\vspace{-0.1cm}
\begin{equation}
\label{eq:theorem-1-2}
\vspace{-0.1cm}
    \mathbf{W}_i = ( \sum\nolimits_{j:(x_j,y_j) \in \mathcal{\hat D}_i} \mathbf{f}_j^\top \mathbf{f}_j +\gamma\mathbf{I})^{-1} \sum\nolimits_{j:(x_j,y_j) \in \mathcal{\hat D}_i} \mathbf{f}_j^\top \mathbf{y}_j.
    \nonumber
\end{equation}

\noindent \textbf{\textit{Proof.}} The derivation is based on the least-squares method with $\ell_{2}$ regularization.
See Appendix A for details.

We then prove that the model $\mathbf{W}_{i}$, updated recursively via \eqref{eq:subsection-3.3-2-Unlearning}, is equivalent to the re-trained model obtained from $\mathcal{\hat D}_i$ in {Theorem 1}, establishing the exact unlearning of ACU.

\noindent\textbf{Theorem 1:} Consider the recursive updating formula:
\vspace{-0.1cm}
\begin{equation}
\vspace{-0.1cm}
\label{eq:Theorem-3-1}
\mathbf{ W}_{i} =   (\mathbf{I} + \mathbf{T}_i  \sum\nolimits_{\mathcal{\check D}_{i}} \mathbf{f}_j^\top \mathbf{f}_j)\mathbf{W}_{i-1} - \mathbf{T}_i \sum\nolimits_{ \mathcal{\check D}_{i}} \mathbf{f}_j^\top \mathbf{y}_j.
\end{equation}
The resulting model $\mathbf{W}_i$ is exactly equivalent to the closed-form solution of the optimization problem \eqref{eq:Motivation-1}, representing the optimal re-trained model directly obtained from $\mathcal{\hat D}_i$.

\noindent\textbf{\textit{Proof.}} 
The proof involves two main steps: 
(1) inductively establishing the closed-form expression for $\mathbf{T}_i$ utilizing the \textit{Woodbury Matrix Identity}, and (2) substituting this expression for $\mathbf{T}_i$ into \eqref{eq:Theorem-3-1}.
By proving that each unlearned model of our ACU is identical to the re-trained model in the parameter space (deterministically instead of probabilistically), its exact unlearning is evident.
See Appendix A for details.

Meanwhile, we provide a more detailed analysis of the recursive computation for the CL phase in Appendix D and the Class-level Fogetting of our ACU in Appendix E

\begin{table*}[t]
    \centering
    \renewcommand{\arraystretch}{1.3}
    \caption{
    Comparisons of the differences from the re-trained model regarding the model parameters, retained set accuracy, forgetting set accuracy, test set accuracy, and MIA indicator. 
    The {\color{my1}\textbf{bold}} and {\color{my2}\underline{underlined}} results indicate the best and second-best performance, respectively.
    All experiments are conducted three times, and the results are shown as $\mathrm{Mean}_{\pm \mathrm{Standered\;  Deviation}}$.
    }
    \label{table-performance}
    \resizebox{\textwidth}{!}{
        \begin{NiceTabular}{ l| l| c| c c| c c| c c| c c| c c}
        \toprule
         \multirow{2}{*}{Dataset}  & \multirow{2}{*}{Method} & \multirow{2}{*}{Privacy} 
         & \multicolumn{2}{c}{ {\large$\Delta_\text{Params}$} }
         &\multicolumn{2}{c}{ {\large$\Delta_\text{Retain}$} } & \multicolumn{2}{c}{ {\large$\Delta_\text{Forget}$} } & \multicolumn{2}{c}{ {\large$\Delta_\text{Test}$} } & \multicolumn{2}{c}{ {\large$\Delta_\text{MIA}$} } \\
         \cline{4-13}
         &  &  & $K=5$ & $K=25$ & $K=5$ & $K=25$ & $K=5$ & $K=25$ & $K=5$ & $K=25$ & $K=5$ & $K=25$ \\

        \cline{1-13}
        \multirow{8}{*}{CIFAR-10}  
        &  Finetuning  & \XSolidBrush & ${33.65}_{\pm 0.06}$ & ${35.45}_{\pm 0.95}$ & ${6.61}_{\pm 0.09}$ & ${6.09}_{\pm 4.09}$ & ${1.03}_{\pm 0.50}$ & ${4.19}_{\pm 1.53}$ & ${2.26}_{\pm 0.70}$ & ${4.20}_{\pm 1.44}$ & ${0.02}_{\pm 0.01}$ & ${0.04}_{\pm 0.02}$\\
        &  L1-Spare  & \XSolidBrush &  {\color{my2}$\underline{{24.14}_{\pm 0.01}}$} & {\color{my2}$\underline{{25.39}_{\pm 0.17}}$} & ${16.79}_{\pm 0.81}$ & ${15.16}_{\pm 3.08}$ & ${9.88}_{\pm 0.74}$ & ${10.26}_{\pm 1.00}$ & ${10.09}_{\pm 0.37}$ & ${10.50}_{\pm 1.53}$ & ${0.49}_{\pm 0.28}$ & ${0.10}_{\pm 0.02}$\\

        &  SCRUB  & \XSolidBrush &  ${28.12}_{\pm 0.00}$ & ${34.34}_{\pm 3.56}$ & {\color{my2}$\underline{{1.17}_{\pm 0.09}}$} & ${15.42}_{\pm 5.23}$ & ${8.48}_{\pm 0.49}$ & ${6.85}_{\pm 4.54}$ & {\color{my2}$\underline{{0.07}_{\pm 0.05}}$} & $8.19_{\pm 4.01}$ & ${0.09}_{\pm 0.01}$ & ${0.07}_{\pm 0.05}$\\
        &  WoodFisher  & \XSolidBrush &  ${28.30}_{\pm 0.00}$ & ${57.12}_{\pm 0.35}$ & ${1.57}_{\pm 0.01}$ & ${89.58}_{\pm 0.07}$ & ${10.02}_{\pm 0.19}$ & ${79.23}_{\pm 0.15}$ & ${0.51}_{\pm 0.04}$ & ${79.23}_{\pm 0.00}$ & ${0.09}_{\pm 0.00}$ & {\color{my2}$\underline{{0.01}_{\pm 0.00}}$}\\
        &  RandomLabel  & \XSolidBrush &  ${27.45}_{\pm 0.01}$ & ${26.75}_{\pm 0.90}$ & ${2.21}_{\pm 0.32}$ & {\color{my2}$\underline{{3.26}_{\pm 2.11}}$} & {\color{my2}$\underline{{0.92}_{\pm 0.51}}$} & {\color{my2}$\underline{{2.47}_{\pm 1.02}}$} & ${2.40}_{\pm 0.39}$ & {\color{my2}$\underline{{3.89}_{\pm 0.26}}$} & {\color{my2}$\underline{{0.01}_{\pm 0.00}}$} & ${0.03}_{\pm 0.00}$\\
        &  NegGrad+  & \XSolidBrush &  ${28.63}_{\pm 0.01}$ & ${55.50}_{\pm 9.60}$ & ${13.21}_{\pm 0.79}$ & ${82.32}_{\pm 3.61}$ & ${9.73}_{\pm 1.30}$ & ${71.85}_{\pm 3.67}$ & ${11.51}_{\pm 0.62}$ & ${70.91}_{\pm 3.89}$ & ${0.09}_{\pm 0.01}$ & ${0.07}_{\pm 0.04}$\\
        &  NegGrad  & \CheckmarkBold &  ${28.30}_{\pm 0.00}$ & ${28.31}_{\pm 0.00}$ & ${6.06}_{\pm 0.02}$ & ${24.85}_{\pm 1.30}$ & ${1.57}_{\pm 0.51}$ & ${14.82}_{\pm 1.52}$ & ${6.55}_{\pm 0.03}$ & ${20.20}_{\pm 0.66}$ & ${0.02}_{\pm 0.00}$ & ${0.15}_{\pm 0.01}$\\
        & ACU & \CheckmarkBold & {\color{my1}$\textbf{00.00}_{\pm 0.00}$} & ${\color{my1}\textbf{00.00}_{\pm 0.00}}$ & {\color{my1}$\textbf{00.00}_{\pm 0.00}$} & {\color{my1}$\textbf{00.00}_{\pm 0.00}$} & {\color{my1}$\textbf{00.00}_{\pm 0.00}$} & {\color{my1}$\textbf{00.00}_{\pm 0.00}$} & {\color{my1}$\textbf{00.00}_{\pm 0.00}$} & {\color{my1}$\textbf{00.00}_{\pm 0.00}$} & {\color{my1}$\textbf{00.00}_{\pm 0.00}$} & {\color{my1}$\textbf{00.00}_{\pm 0.00}$}\\
        \cline{1-13}
        \multirow{8}{*}{CIFAR-100}  
        &  Finetuning  & \XSolidBrush & ${58.92}_{\pm 0.03}$ & ${61.80}_{\pm 2.94}$ & ${12.03}_{\pm 0.23}$ & ${15.55}_{\pm 2.81}$ & {\color{my2}$\underline{{1.27}_{\pm 0.57}}$} & {\color{my2}$\underline{{4.03}_{\pm 4.36}}$} & {\color{my2}$\underline{{0.30}_{\pm 0.09}}$} & ${4.72}_{\pm 4.09}$ & {\color{my2}$\underline{{0.02}_{\pm 0.01}}$} & ${0.04}_{\pm 0.04}$\\
        &  L1-Spare  & \XSolidBrush &  {\color{my2}$\underline{{44.93}_{\pm 0.00}}$} & {\color{my2}$\underline{{45.64}_{\pm 0.72}}$} & ${29.36}_{\pm 1.84}$ & ${30.17}_{\pm 9.19}$ & ${11.85}_{\pm 1.03}$ & ${14.86}_{\pm 0.71}$ & ${13.55}_{\pm 1.98}$ & ${15.33}_{\pm 1.69}$ & ${0.12}_{\pm 0.02}$ & ${0.15}_{\pm 0.01}$\\
        &  SCRUB  & \XSolidBrush &  ${53.72}_{\pm 0.00}$ & ${54.59}_{\pm 1.94}$ & ${0.98}_{\pm 0.19}$ & {\color{my2}$\underline{{7.60}_{\pm 0.97}}$} & ${20.22}_{\pm 0.83}$ & ${10.65}_{\pm 3.74}$ & ${2.47}_{\pm 0.03}$ & {\color{my2}$\underline{{2.76}_{\pm 3.50}}$} & ${0.20}_{\pm 0.00}$ & ${0.11}_{\pm 0.04}$\\
        &  WoodFisher  & \XSolidBrush &  ${54.00}_{\pm 0.00}$ & ${54.01}_{\pm 0.00}$ & ${1.38}_{\pm 0.15}$ & ${16.04}_{\pm 9.15}$ & ${22.7}_{\pm 1.09}$ & ${11.87}_{\pm 7.37}$ & ${3.06}_{\pm 0.10}$ & ${4.50}_{\pm 5.50}$ & ${0.23}_{\pm 0.00}$ & ${0.12}_{\pm 0.07}$\\
        &  RandomLabel  & \XSolidBrush &  ${53.77}_{\pm 0.01}$ & ${53.76}_{\pm 3.00}$ & ${7.16}_{\pm 0.29}$ & ${11.89}_{\pm 0.38}$ & ${6.42}_{\pm 0.51}$ & ${11.72}_{\pm 1.12}$ & ${2.73}_{\pm 0.27}$ & ${5.99}_{\pm 1.07}$ & ${0.63}_{\pm 0.00}$ & {\color{my2}$\underline{{0.02}_{\pm 0.01}}$}\\
        &  NegGrad+  & \XSolidBrush &  ${53.82}_{\pm 0.00}$ & ${54.76}_{\pm 0.60}$ & ${1.09}_{\pm 0.10}$ & ${11.02}_{\pm 7.59}$ & ${9.65}_{\pm 0.67}$ & ${15.07}_{\pm 6.74}$ & ${0.62}_{\pm 0.35}$ & ${9.95}_{\pm 3.09}$ & ${0.10}_{\pm 0.00}$ & ${0.05}_{\pm 0.07}$\\
        &  NegGrad  & \CheckmarkBold &  ${54.00}_{\pm 0.00}$ & ${54.01}_{\pm 0.00}$ & {\color{my2}$\underline{{0.38}_{\pm 0.23}}$} & ${15.53}_{\pm 7.77}$ & ${22.05}_{\pm 0.35}$ & ${12.14}_{\pm 7.64}$ & ${2.31}_{\pm 0.08}$ & ${4.10}_{\pm 4.47}$ & ${0.21}_{\pm 0.00}$ & ${0.12}_{\pm 0.08}$\\
        & ACU & \CheckmarkBold & {\color{my1}$\textbf{00.00}_{\pm 0.00}$} & {\color{my1}$\textbf{00.00}_{\pm 0.00}$} & {\color{my1}$\textbf{00.00}_{\pm 0.00}$} & {\color{my1}$\textbf{00.00}_{\pm 0.00}$} & {\color{my1}$\textbf{00.00}_{\pm 0.00}$} & {\color{my1}$\textbf{00.00}_{\pm 0.00}$} & {\color{my1}$\textbf{00.00}_{\pm 0.00}$} & {\color{my1}$\textbf{00.00}_{\pm 0.00}$} & {\color{my1}$\textbf{00.00}_{\pm 0.00}$} & {\color{my1}$\textbf{00.00}_{\pm 0.00}$}\\
        \bottomrule
        \end{NiceTabular}
     }
\end{table*}

Furthermore, we provide comprehensive efficiency analyses of ACU in Appendix B.
Specifically, for each unlearning request $i$, the computational and storage overheads are $O({d_\mathrm{F}}^3 + |\mathcal{\check D}_i|{d_\mathrm{F}}^2 + |\mathcal{\check D}_i|^2d_\mathrm{F} + |\mathcal{\check D}_i|^3)$ and $O({d_\mathrm{F}}^2 + d_\mathrm{C}d_\mathrm{F})$. 
Notably, the cubic terms in the computational overhead stem from matrix multiplications, for which we employ the loose bound $O(n^3)$.
In fact, recent advances have lowered this complexity to $O(n^{2.373})$, indicating that ACU is even more efficient than our foregoing claim suggests.

Last but not least, we also provide detailed privacy analyses in Appendix C.
Specifically, we present our ACU's \textit{non-distinguishability} under the \textit{semi-orthogonal transforms}, proving that it is impossible to fully reconstruct the retained set $\mathcal{\hat D}_i$ via reverse engineering only based on the \textit{Knowledge Tracking Matrix} $\mathbf{T}_i$ and the corresponding model $\mathbf{W}_i$.

\section{Experiments}
\label{sec:Experiments}

\subsection{Experimental Setup}
\label{sec:Experimental Setup}

\textbf{Datasets \& Settings.} 
In our experiments, we comprehensively analyze the performance of ACU on two benchmark datasets: CIFAR-10 \cite{dataset_1} and CIFAR-100 \cite{dataset_1}.
Existing PTM-based methods typically finetune the PTM to adapt to the current dataset (also called the base training) before freezing its parameters during the CL phase~\cite{ACIL_1,ACIL_2,ACIL_5,ACIL_L3A}.
Following this, we partition the dataset into three disjoint subsets: one for fine-tuning the PTM, one for CL training, and one for testing.
Subsequently, we select a subset from the CL samples and partition it into 5 to 50 forgetting sets for CU.
More details are given in subsection A of Appendix F.

\textbf{Baselines \& Metrics.} 
We compare ACU with a variety of state-of-the-art unlearning baselines, including Finetuning~\cite{unlearning-8}, L1-Sparsity~\cite{unlearning-9}, NegGrad~\cite{unlearning-11}, NegGrad+~\cite{unlearning-1}, SCRUB~\cite{unlearning-1}, WoodFisher~\cite{unlearning-12}, and RandomLabel~\cite{unlearning-14}. 
Given the prohibitive time overhead of exact unlearning methods in CU (e.g., taking more than 1.5 hours for a single request), we primarily adopt approximate unlearning methods for comparison.
As these baselines are gradient-based, we ensure fairness by deriving their original models through centralized joint training to prevent forgetting, instead of CL, which incidentally improves their performance.
To dispel concerns, subsection ~\ref{Appendix:Experimental-Original Model} also details a parallel experiment where all methods shared the same original model obtained through ACL. Details of these baselines are provided below.
\begin{itemize}
    \item \textbf{Finetune} \cite{unlearning-8}: This method leverages the phenomenon of catastrophic forgetting by simply fine-tuning the model on the retained set, thereby diminishing the influence of the forgetting set.
    \item \textbf{L1-Sparsity} \cite{unlearning-9}: This approach introduces weight sparsity into the unlearning process. The model is fine-tuned on the retained set with an L1 regularization term.
    \item \textbf{NegGrad} \cite{unlearning-11}: Based on the intuitive idea of forgetting by increasing error, this method forces the model to maximize the loss on the forgetting set. However, this may significantly impair the overall model performance.
    \item \textbf{NegGrad+} \cite{unlearning-1}: This method combines the \textbf{Finetune} and \textbf{NegGrad}, aiming to balance forgetting and performance retention by simultaneously maximizing the loss on the forget set while minimizing it on the retain set.
    \item \textbf{SCRUB} \cite{unlearning-1}: Extending the \textbf{NegGrad+} framework, this method recasts unlearning as a student-teacher distillation problem. It minimizes the distributional discrepancy between the student and teacher models on the retained set, while maximizing it on the forgetting set.
    \item \textbf{WoodFisher} \cite{unlearning-12}:  This aims to achieve $(\epsilon, \delta)$-forgetting by estimating the influence of individual parameters and removing their contributions to the forgetting data.
    \item \textbf{RandomLabel} \cite{unlearning-14}: This method randomly relabels the forgetting set and then finetunes the model jointly on the relabeled forgetting set and the retained set.
\end{itemize}

\begin{figure*}[t] 
\centering 
	\begin{minipage}[t]{0.329\linewidth}
		\centering
		\subfloat[Retained set accuracy]{ %
              \centering  
               \label{fig:continual-retain}
               \includegraphics[width=1.0\textwidth]{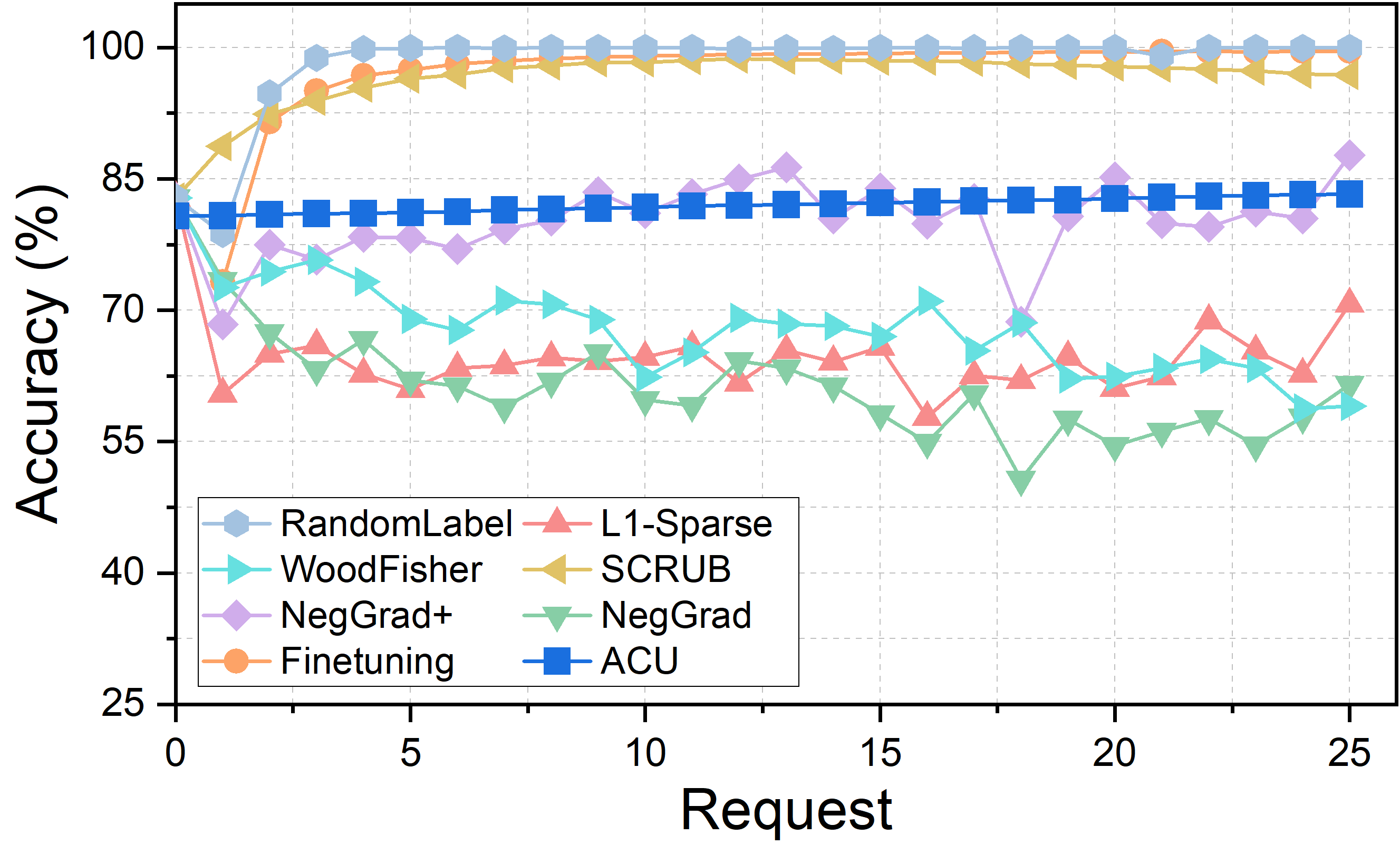}
             } 
	\end{minipage}
	\hfill 
	\begin{minipage}[t]{0.329\linewidth}
		\centering
		\subfloat[Test set accuracy]{ %
              \centering  
               \label{fig:continual-test}
               \includegraphics[width=1.0\textwidth]{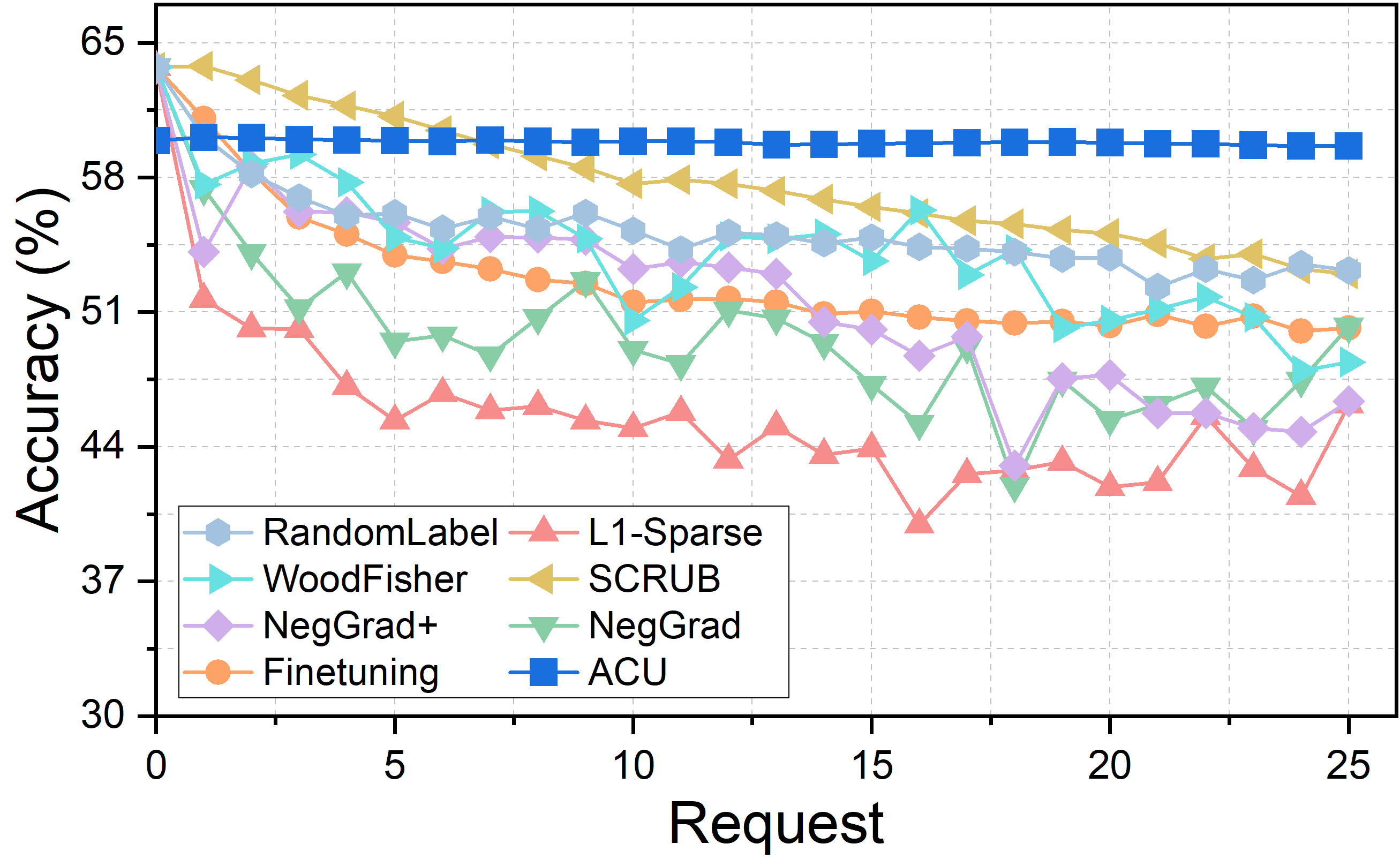}
             } 
	\end{minipage}
        \hfill 
	\begin{minipage}[t]{0.329\linewidth}
		\centering
		\subfloat[Cumulative time cost]{ %
              \centering  
               \label{fig:continual-time}
               \includegraphics[width=1.0\textwidth]{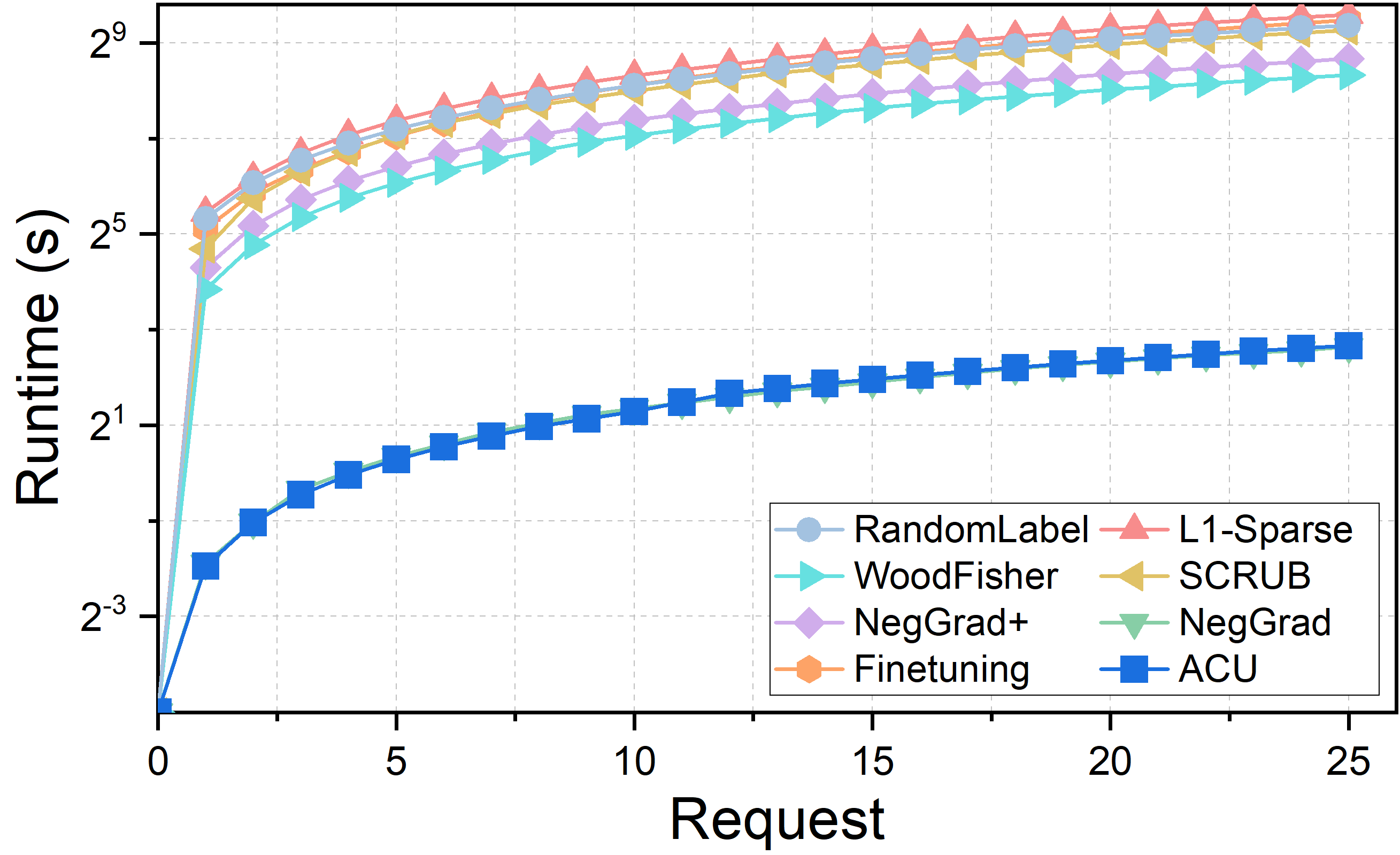}
             } 
	\end{minipage}
 \caption{The dynamics analysis for a total of 25 CU requests on the CIFAR-100 dataset.}  %
 \label{fig:continual-unlearning-3}
 \vspace{0.3cm}
 \begin{minipage}[t]{0.329\linewidth}
		\centering
		\subfloat[Retained set accuracy]{ %
              \centering  
               \label{fig:continual-retain}
               \includegraphics[width=1.0\textwidth]{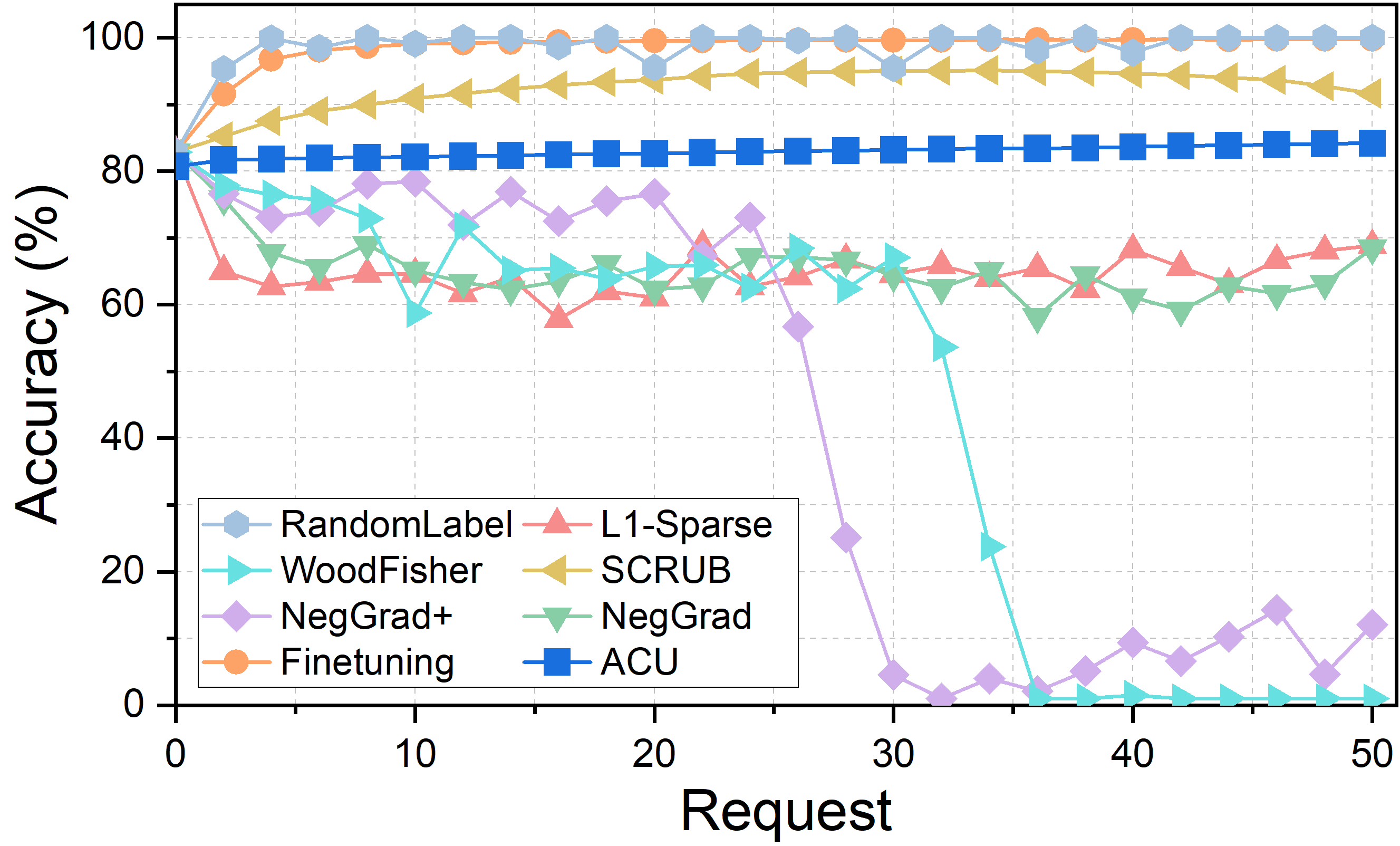}
             } 
	\end{minipage}
	\hfill 
	\begin{minipage}[t]{0.329\linewidth}
		\centering
		\subfloat[Test set accuracy]{ %
              \centering  
               \label{fig:continual-test}
               \includegraphics[width=1.0\textwidth]{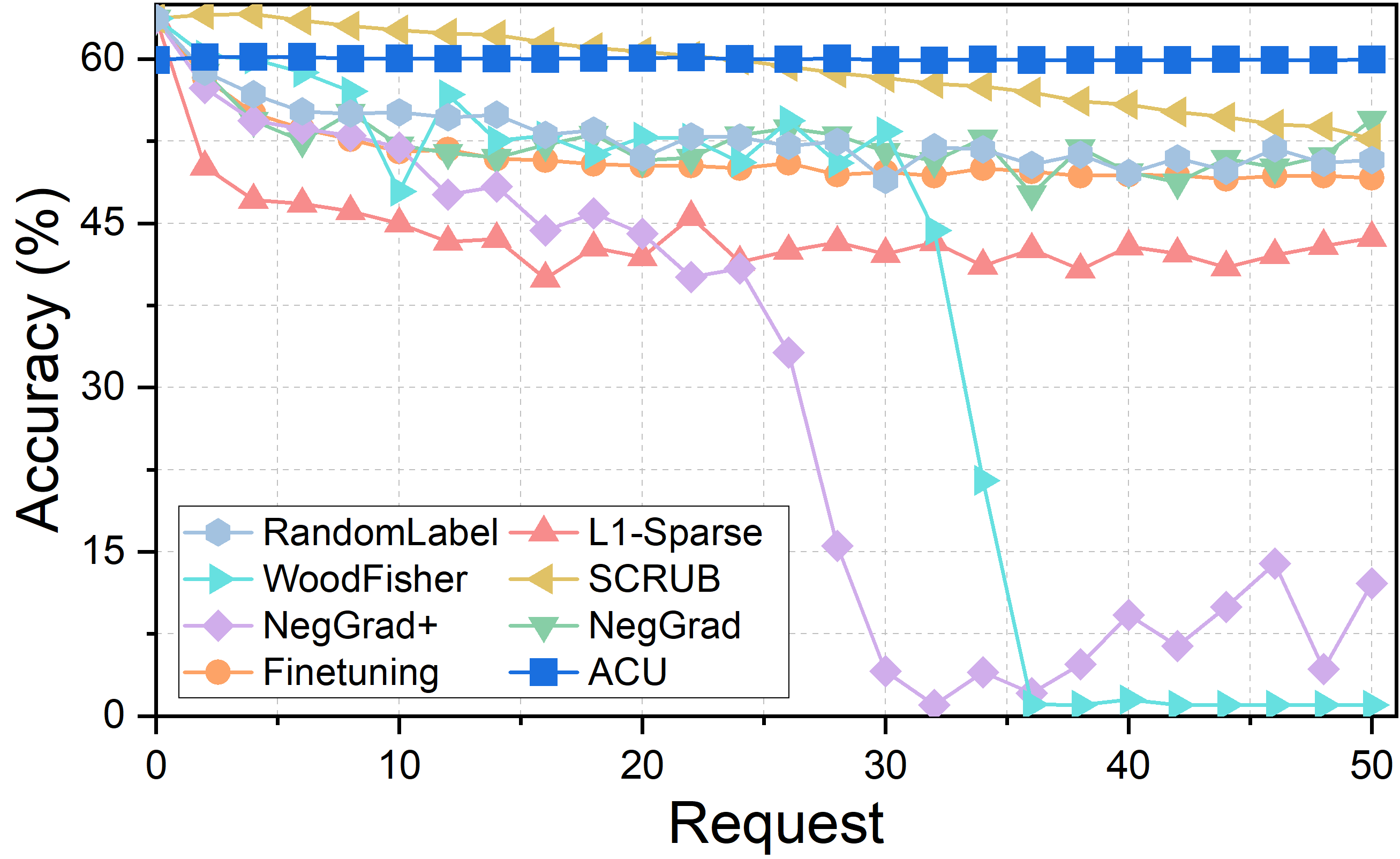}
             } 
	\end{minipage}
        \hfill 
	\begin{minipage}[t]{0.329\linewidth}
		\centering
		\subfloat[Cumulative time cost]{ %
              \centering  
               \label{fig:continual-time}
               \includegraphics[width=1.0\textwidth]{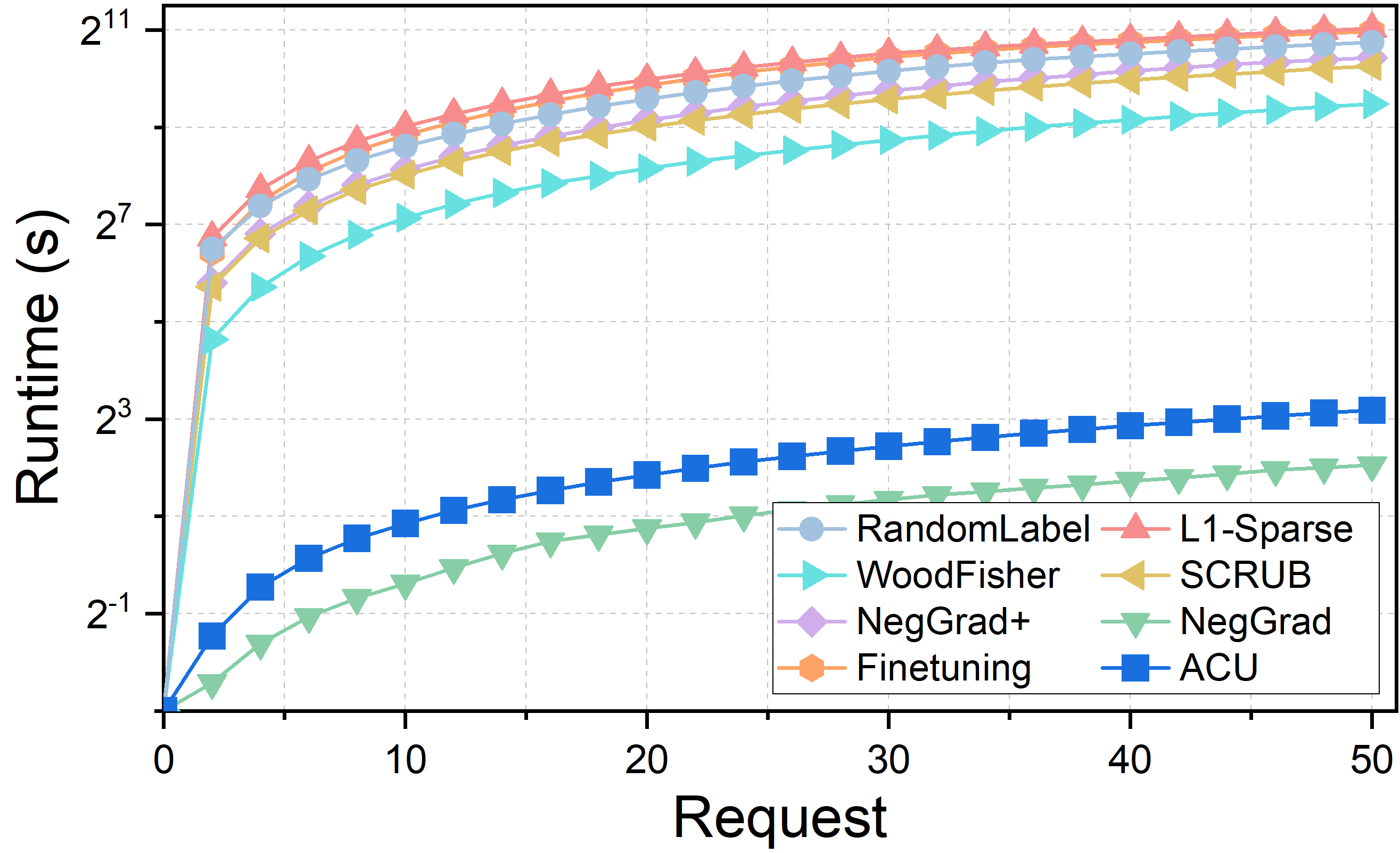}
             } 
	\end{minipage}
    \vspace{-0.1cm}
 \caption{The dynamics analysis for a total of  50 CU requests on the CIFAR-100 dataset.}  %
 \label{fig:continual-unlearning-4}
    \vspace{-0.1cm}
\end{figure*} 

Furthermore, we assess each method by quantifying its differences from the optimal retrained model in terms of model parameters, average accuracy, and vulnerability to Membership Inference Attacks (MIA) \cite{MIA}.
Moreover, cumulative runtime is used to evaluate efficiency.
We provide more details of baselines and metrics in Appendix F.

\subsection{Overall Analysis on Unlearning Effectiveness}

As demonstrated in Table~\ref{table-performance}, we first measure the differences of all methods from the re-trained model using several metrics: model parameters gap $\Delta_\text{Params}$ ($\text{L}_\text{2}$ Norm), retained set accuracy gap $\Delta_\text{Retain}$, forgetting set accuracy gap $\Delta_\text{Forget}$, test set accuracy gap $\Delta_\text{Test}$, and the MIA indicator gap $\Delta_\text{MIA}$, to comprehensively compare their unlearning capabilities.
Thanks to its theoretical guarantee of exact unlearning, our ACU achieves optimal zero values across all metrics, indicating that the unlearned model produced by our ACU is identical to the optimal re-trained model.
In contrast, due to the baselines' inherent reliance on approximate unlearning, they exhibit non-zero values across various metrics.

Notably, the baselines' performance tends to be inconsistent across different metrics.
For example, on the CIFAR-100 dataset, Finetuning achieves favorable results on $\Delta_\text{Test}$ and $\Delta_\text{Forget}$, yet fails to maintain competitive performance on $\Delta_\text{Retain}$.
This phenomenon is reasonable, as approximate unlearning often entails trade-offs: insufficient unlearning may leave residual influence from the data that should have been forgotten, while excessive unlearning may damage unrelated knowledge.
Such imbalances lead to partial effectiveness, i.e., strong performance on certain metrics, but degradation on others.
In comparison, our ACU can achieve exact unlearning without any trade-off, enabling it to maintain optimal performance across all metrics.

It is worth emphasizing that most baselines rely on revisiting the historical data from the retained set for fine-tuning during the CU phase, which essentially constitutes a form of \textit{cheating}.
This requirement is fundamentally incompatible with practical CU scenarios, in accordance with the CL philosophy that online task data is discarded after training.
The baseline that does not require any access to the historical data is NegGrad~\cite{unlearning-11}, which updates the model by applying reverse gradient descent on the small forgetting set.
However, due to the lack of access to the retained data, NegGrad exhibits relatively poor performance across various evaluation metrics.
In contrast, ACU maintains its superior performance without any access to historical data from the retained set, further highlighting its advantages of the gradient-free updates.

\subsection{Dynamics Analysis on Fidelity and Efficiency}

\begin{table*}[t]
    \centering
    \renewcommand{\arraystretch}{1.1}
    \caption{
     Comparisons of the differences from the re-trained model using the unified ACL-based original model regarding the model parameters, retained set accuracy, forgetting set accuracy, test set accuracy, and MIA indicator. 
    The {\color{my1}\textbf{bold}} and {\color{my2}\underline{underlined}} results indicate the best and second-best performance, respectively.
    }
    \label{table-acl}
    \resizebox{\textwidth}{!}{
        \begin{NiceTabular}{ l| l| c| c c| c c| c c| c c| c c}
        \toprule
         \multirow{2}{*}{Dataset}  & \multirow{2}{*}{Method} & \multirow{2}{*}{Privacy} 
         & \multicolumn{2}{c}{ {\large$\Delta_\text{Params}$} }
         &\multicolumn{2}{c}{ {\large$\Delta_\text{Retain}$} } & \multicolumn{2}{c}{ {\large$\Delta_\text{Forget}$} } & \multicolumn{2}{c}{ {\large$\Delta_\text{Test}$} } & \multicolumn{2}{c}{ {\large$\Delta_\text{MIA}$} } \\
         \cline{4-13}
         &  &  & $K=5$ & $K=25$ & $K=5$ & $K=25$ & $K=5$ & $K=25$ & $K=5$ & $K=25$ & $K=5$ & $K=25$ \\

        \cline{1-13}
        \multirow{8}{*}{CIFAR-10}  
        &  Finetuning  & \XSolidBrush & \color{my2}\underline{28.26} & \color{my2}\underline{28.34} & \color{my2}\underline{6.34} & \color{my2}\underline{6.28} & \color{my2}\underline{4.24} & 4.06 & 0.31 & \color{my2}\underline{0.22} & 0.39 & 0.38 \\
        &  L1-Spare  & \XSolidBrush &  28.45 & 28.48 & 6.63 & 6.66 & 4.59 & 4.59 & 0.55 & 0.43 & 0.38 & 0.38 \\

        &  SCRUB  & \XSolidBrush &  29.53 & 29.36 & 6.57 & 6.49 & 4.70 & 4.22 & \color{my2}\underline{0.27} & 0.27 & 0.38 & 0.39 \\
        &  WoodFisher  & \XSolidBrush & 29.60 & 29.60 & 79.69 & 80.05 & 77.20 & 77.67 & 74.39 & 74.77 & 0.38 & 0.38 \\
        &  RandomLabel  & \XSolidBrush & 29.51 & 29.21 & 6.76 & 6.28 & 4.57 & \color{my2}\underline{3.92} & 0.52 & 0.42 & 0.38 & 0.39 \\
        &  NegGrad+  & \XSolidBrush &  30.00 & 251.58 & 22.03 & 40.29 & 20.14 & 37.50 & 19.24 & 37.77 & \color{my2}\underline{0.22} & \color{my2}\underline{0.06} \\
        &  NegGrad  & \CheckmarkBold &  89.28 & 65.83 & 83.39 & 83.39 & 81.06 & 81.06 & 77.59 & 77.59 & 0.42 & 0.42 \\
        & ACU & \CheckmarkBold & \color{my1}\textbf{0.00} & \color{my1}\textbf{0.00} & \color{my1}\textbf{0.00} & \color{my1}\textbf{0.00} & \color{my1}\textbf{0.00} & \color{my1}\textbf{0.00} & \color{my1}\textbf{0.00} & \color{my1}\textbf{0.00} & \color{my1}\textbf{0.00} & \color{my1}\textbf{0.00} \\
        \cline{1-13}
        \multirow{8}{*}{CIFAR-100}  
        &  Finetuning  & \XSolidBrush & 28.51 & 35.04 & \color{my2}\underline{24.17} & 24.11 & \color{my2}\underline{13.08} & \color{my2}\underline{12.77} & \color{my2}\underline{2.00} & \color{my2}\underline{1.74} & \color{my2}\underline{0.14} & 0.14 \\
        &  L1-Spare  & \XSolidBrush & 13.60 & 13.89 & 27.15 & 26.07 & 14.79 & 14.51 & 3.60 & 2.83 & 0.26 & 0.26 \\
        &  SCRUB  & \XSolidBrush & 3.54 & 4.61 & 25.40 & 24.54 & 13.99 & 13.54 & 2.64 & 2.21 & 0.27 & 0.13 \\
        &  WoodFisher  & \XSolidBrush & \color{my2}\underline{2.51} & \color{my2}\underline{2.51} & 83.49 & 83.43 & 69.81 & 69.76 & 59.40 & 59.40 & 0.29 & 0.29 \\
        &  RandomLabel  & \XSolidBrush & 4.42 & 5.58 & 24.78 & \color{my2}\underline{23.99} & 13.29 & 13.23 & 2.31 & 1.99 & 0.28 & \color{my2}\underline{0.13} \\
        &  NegGrad+  & \XSolidBrush & 4.43 & 10.82 & 25.44 & 30.32 & 15.33 & 19.79 & 3.31 & 7.98 & 0.15 & 0.19 \\
        &  NegGrad  & \CheckmarkBold & 11284.60 & 9152.29 & 83.07 & 83.07 & 69.38 & 69.38 & 59.01 & 59.01 & 0.29 & 0.29 \\
        & ACU & \CheckmarkBold & \color{my1}\textbf{0.00} & \color{my1}\textbf{0.00} & \color{my1}\textbf{0.00} & \color{my1}\textbf{0.00} & \color{my1}\textbf{0.00} & \color{my1}\textbf{0.00} & \color{my1}\textbf{0.00} & \color{my1}\textbf{0.00} & \color{my1}\textbf{0.00} & \color{my1}\textbf{0.00}\\
        \bottomrule
        \end{NiceTabular}
     }
\end{table*}

Subsequently, we investigate the fidelity and efficiency of all methods when processing dynamically arriving CU requests.
Specifically, we select 10,000 samples from the CL samples in CIFAR-100 and partition them into 25 or 50 disjoint CU requests.
We then record the performance of each method across varying numbers of CU requests in terms of (a) retained set accuracy, (b) test set accuracy, and (c) cumulative time cost to assess their fidelity and efficiency, as illustrated in Figures~\ref{fig:continual-unlearning-3}--\ref{fig:continual-unlearning-4}.
More results are presented in Figures F-1 to F-2, as well as Tables F-I to F-VI within Appendix F.

As shown in Figure~\ref{fig:continual-unlearning-3}(a) and Table F-I, the retained set accuracy of NegGrad, L1-Sparse, and WoodFisher undergoes catastrophic degradation as CU requests accumulate.
This degradation stems from the fact that these methods, during their approximate CU processes, inevitably impair the knowledge originally preserved in the retained set.
In contrast, the accuracy of ACU on the retained set remains stable and gradually improves.
This stems from ACU's ability to achieve exact forgetting while simultaneously amplifying the influence of the retained data.
Only four other baseline methods manage to preserve or slightly improve their retained set accuracy, primarily due to fine-tuning with access to the retained data.
Notably, the preservation of retained set accuracy by these baselines is achieved by violating the constraint of not accessing historical data during the CU phase, which can be considered a form of \textit{cheating}.
Moreover, such preservation of baselines on the retained set does not necessarily mean genuine retention of the non-forgotten knowledge, but rather reflects mere overfitting on the retained set due to \textit{fine-tuning}.

Figure~\ref{fig:continual-unlearning-3}(b) and Table F-II provide further evidence that all the baselines exhibit a progressive and substantial decline in test set accuracy as CU requests accumulate, despite some of them achieving relatively good retained set accuracy.
It indicates that the seemingly well-preserved retained set accuracy is merely superficial, resulting from overfitting to the retained set, while the test set accuracy is severely compromised.
These results underscore that baseline methods suffer from a catastrophic erosion of model fidelity due to the unintended forgetting of knowledge that should have been retained.
In contrast, ACU consistently achieves high and stable accuracy on both the retained and test sets.

As shown in Figure~\ref{fig:continual-unlearning-3}(c) and Table F-III, except for NegGrad, all other baselines incur a significantly higher total time cost than our ACU (approximately 50 to 125 times), even though these baselines are already considered relatively efficient among existing methods.
The efficiency advantage of NegGrad stems from its simplicity, as it updates the model solely by applying reverse gradient descent on the small forgetting set.
However, its simplicity compromises its model fidelity, as previously discussed.
For reference, we also briefly discuss the impracticality of re-training from an efficiency perspective in CU.
Specifically, re-training the model from scratch would take approximately 1.5 hours per request (5400 seconds), resulting in a total of over 135,000 seconds for 25 requests (more than 10,000× slower than our proposed ACU).
Thus, the inefficiency renders re-training entirely infeasible for real-world applications.

Subsequently, we extend our analysis to a more challenging scenario by doubling the CU requests from 25 to 50, as presented in Figure~\ref{fig:continual-unlearning-4} and Tables F-IV to F-VI.
From these results, we observe a notably catastrophic degradation in model fidelity for WoodFisher and NegGrad+, which was not observed in Figure~\ref{fig:continual-unlearning-3}. 
This observation indicates that these baselines are sensitive to the number of CU requests and may experience severe fidelity collapse when handling frequent requests.
Meanwhile, the cumulative time cost of our ACU increases only from 12.68 seconds at 25 requests to 18.16 seconds at 50 requests, representing just a 1.43× increase.
By comparison, for instance, NegGrad+'s time cost rises sharply from 815.13 seconds to 2769.09 seconds, a surge of up to 3.40×.
These results highlight that our ACU not only achieves superior efficiency in both scenarios but also shows greater robustness to the number of CU requests.

\subsection{Unified ACL-based Original Model Comparison}
\label{Appendix:Experimental-Original Model}

\begin{figure*}[t] 
\centering 
	\begin{minipage}[t]{0.329\linewidth}
		\centering
		\subfloat[Retained set accuracy]{ %
              \centering  
               \label{fig:continual-retain}
               \includegraphics[width=1.0\textwidth]{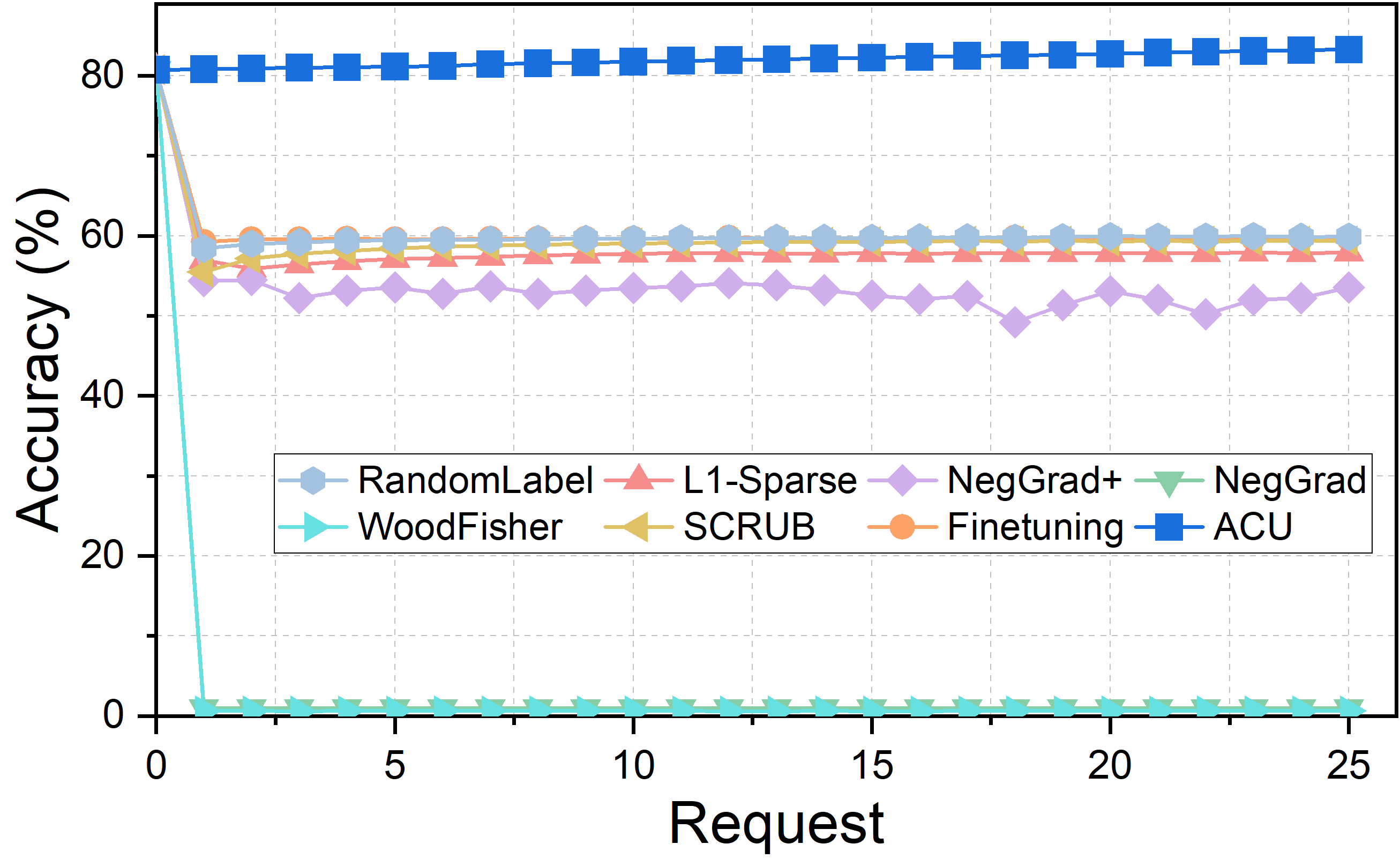}
             } 
	\end{minipage}
	\hfill 
	\begin{minipage}[t]{0.329\linewidth}
		\centering
		\subfloat[Test set accuracy]{ %
              \centering  
               \label{fig:continual-test}
               \includegraphics[width=1.0\textwidth]{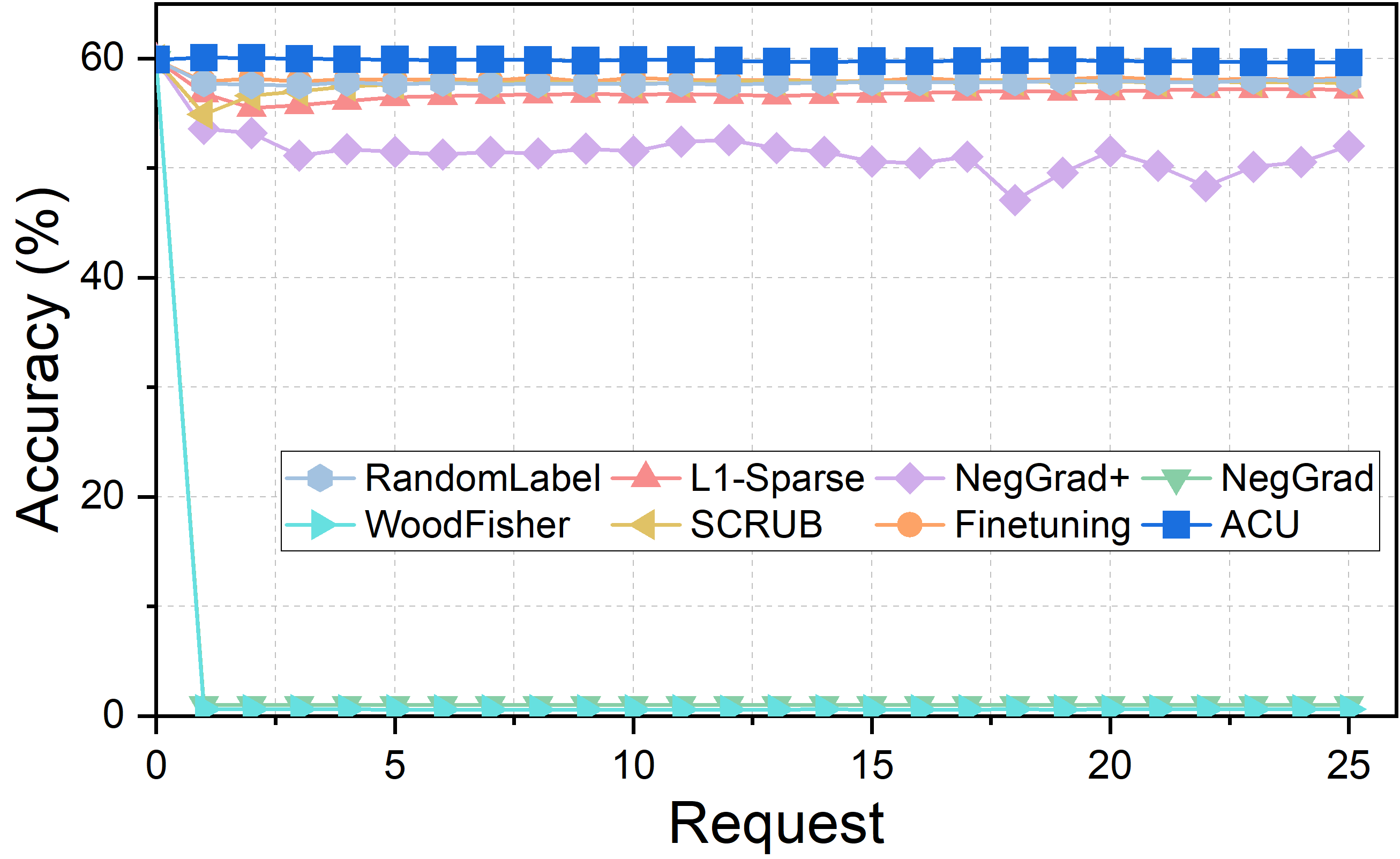}
             } 
	\end{minipage}
        \hfill 
	\begin{minipage}[t]{0.329\linewidth}
		\centering
		\subfloat[Cumulative time cost]{ %
              \centering  
               \label{fig:continual-time}
               \includegraphics[width=1.0\textwidth]{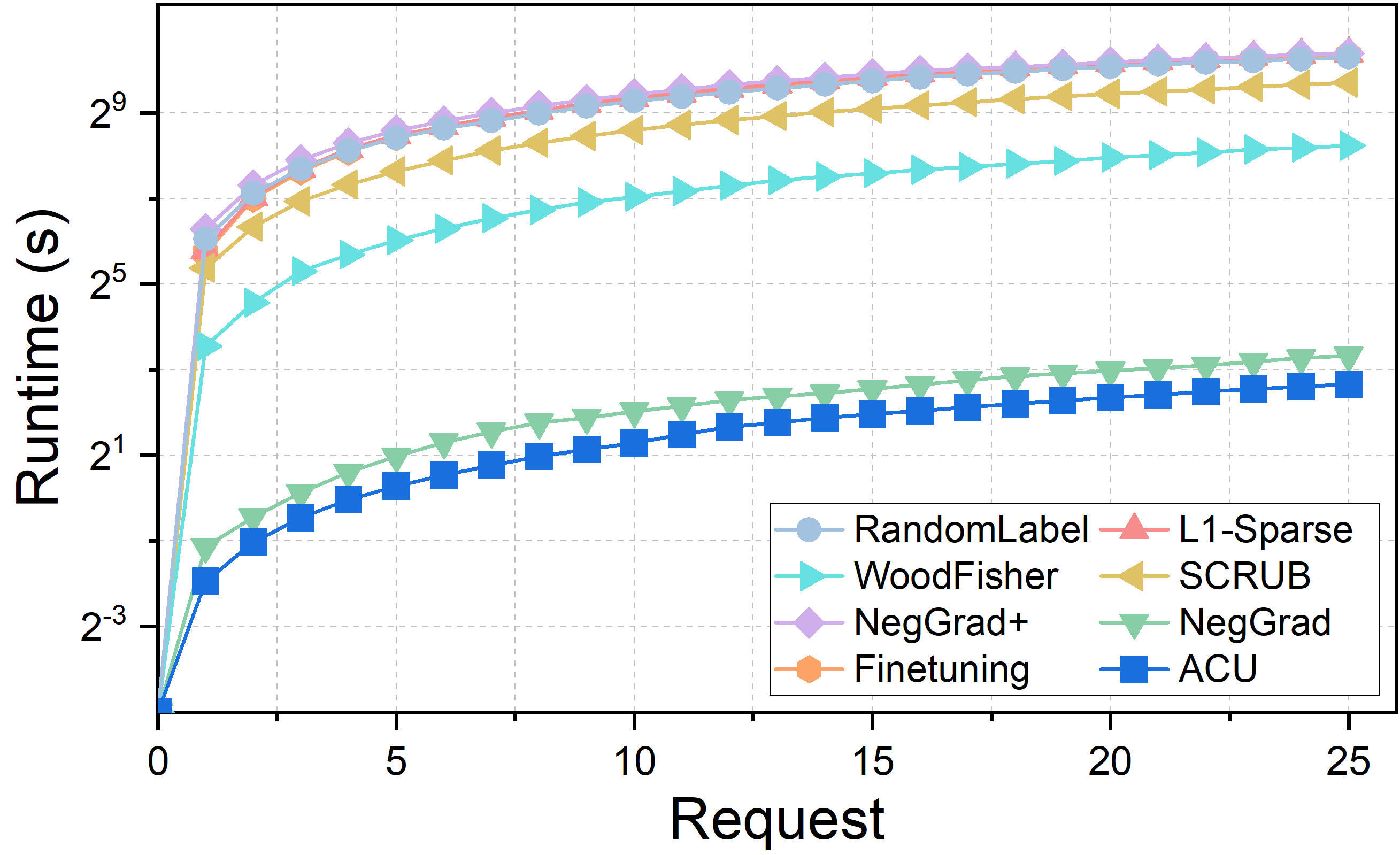}
             } 
	\end{minipage}
    \vspace{-0.1cm}
 \caption{The dynamics analysis for  25 CU requests using Unified ACL-based Original Model on CIFAR-100 dataset.}  %
 \label{fig:app-continual-unlearning-1}
 \vspace{0.3cm}
 \begin{minipage}[t]{0.329\linewidth}
		\centering
		\subfloat[Retained set accuracy]{ %
              \centering  
               \label{fig:continual-retain}
               \includegraphics[width=1.0\textwidth]{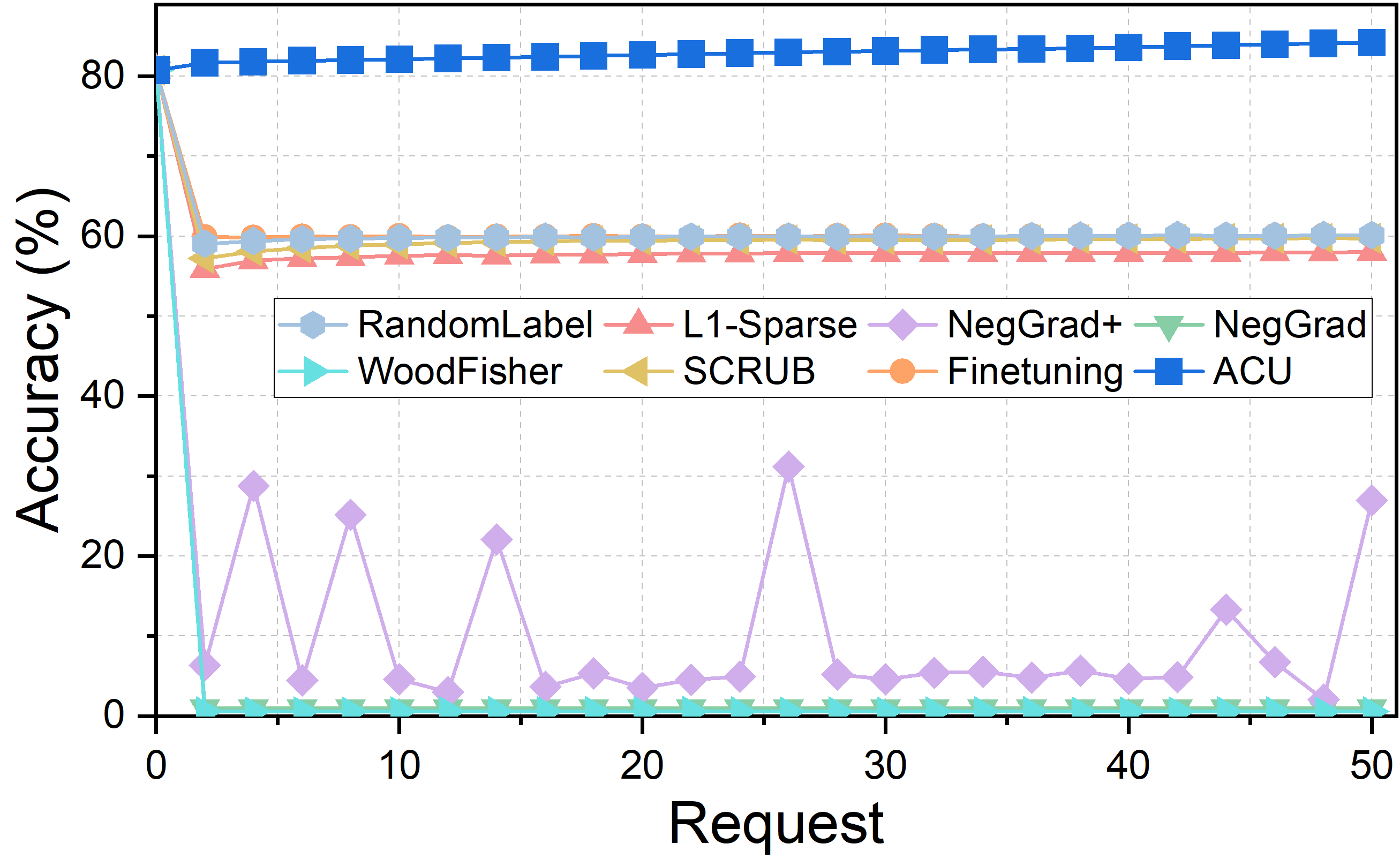}
             } 
	\end{minipage}
	\hfill 
	\begin{minipage}[t]{0.329\linewidth}
		\centering
		\subfloat[Test set accuracy]{ %
              \centering  
               \label{fig:continual-test}
               \includegraphics[width=1.0\textwidth]{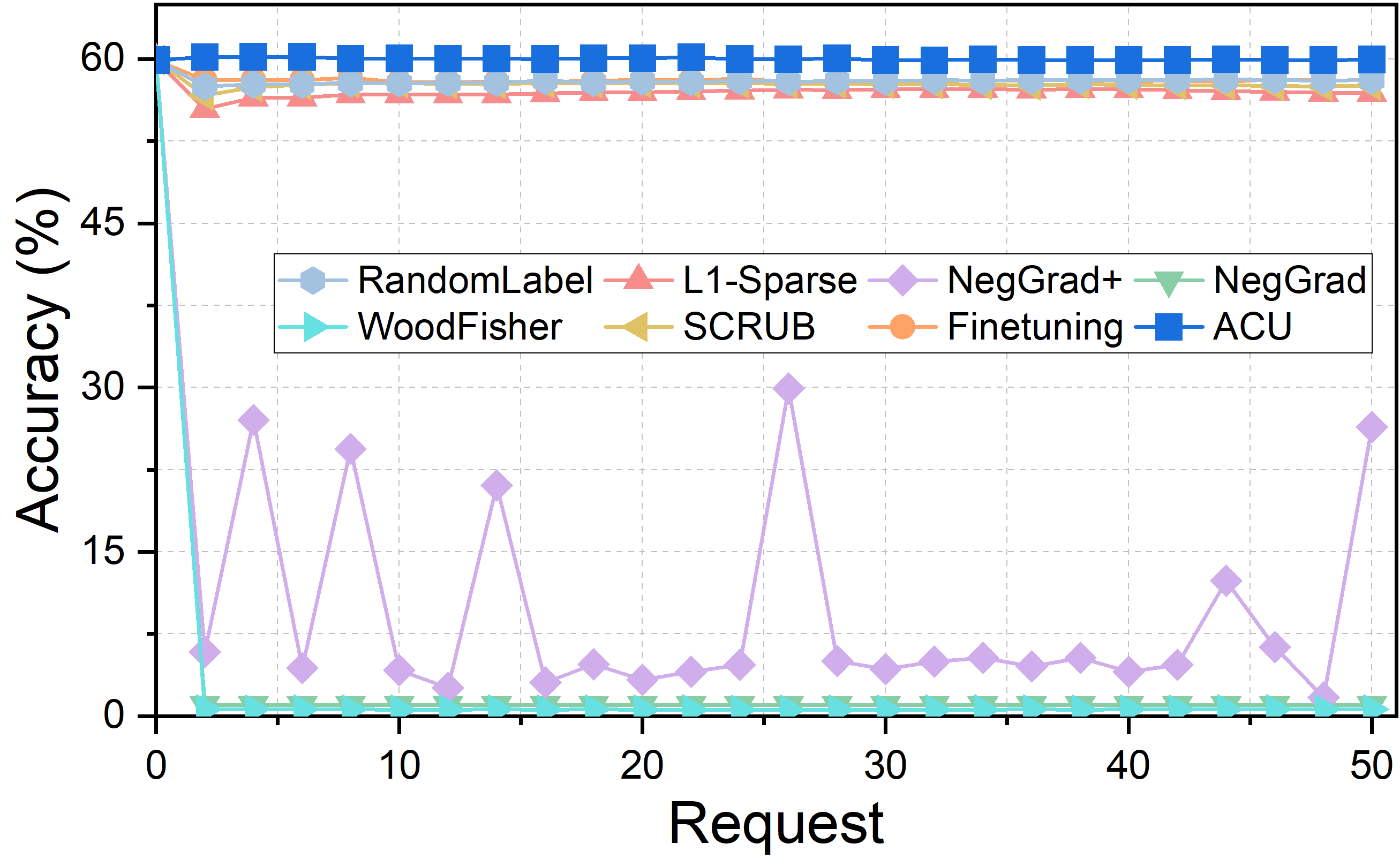}
             } 
	\end{minipage}
        \hfill 
	\begin{minipage}[t]{0.329\linewidth}
		\centering
		\subfloat[Cumulative time cost]{ %
              \centering  
               \label{fig:continual-time}
               \includegraphics[width=1.0\textwidth]{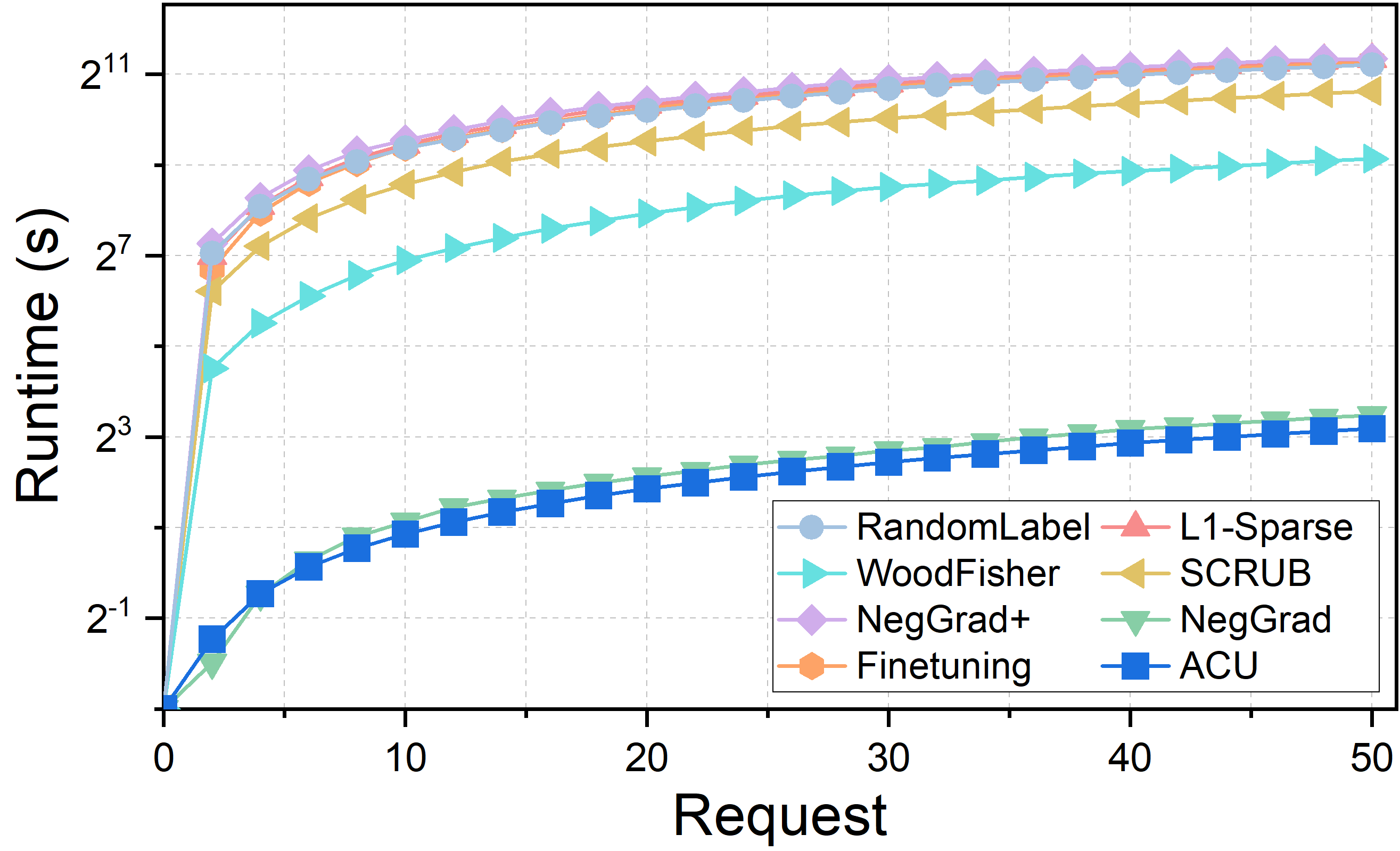}
             } 
	\end{minipage}
    \vspace{-0.1cm}
 \caption{The dynamics analysis for  50 CU requests using Unified ACL-based Original Model on CIFAR-100 dataset.}  %
 \label{fig:app-continual-unlearning-2}
    \vspace{-0.35cm}
\end{figure*}

Here, we further compare unlearning performance by initializing all methods with a unified ACL-based original model as the starting state for the CU phase.
Specifically, all methods are initialized with the same frozen PTM and the same analytic classifier obtained through ACL.
Upon receiving sequential CU requests, all methods perform unlearning by updating only the analytic classifier.
The results are shown in Table~\ref{table-acl} and Figures~\ref{fig:app-continual-unlearning-1}--\ref{fig:app-continual-unlearning-2}. The performance of all baselines on both the retained and test sets converges rapidly and significantly lags behind that of our ACU.
This performance deficit arises because these baselines are fundamentally designed for gradient-based learning, not for analytic learning.
Consequently, applying gradient-based updates to the analytic classifier constitutes a significant methodological mismatch, causing the models to rapidly converge to a suboptimal state.
Furthermore, this incompatibility also led to the collapse of several baselines.

\subsection{Analyses for Class-level Unlearning}
\label{Appendix:Experimental-Class-level}
Here, we present a detailed evaluation of our ACU when subjected to class-level unlearning requests.
The specific experimental results are displayed in Figures F-3 to F-6 in Appendix F.
A key observation is the gradual performance improvement of our ACU on the retained set as the unlearning process progresses.
This trend is primarily due to our ACU judiciously amplifying the influence of the retained set during model updates.
In addition, we note a marginal decline in the performance of our ACU on the test set as the number of unlearning requests steadily increases.
This slight degradation is fundamentally a consequence of the gradual attrition of training samples, which naturally diminishes the model's overall generalization capability.
What's more, the cumulative time cost for our ACU is remarkably low, taking less than 15 seconds to unlearn up to 25 classes and less than 40 seconds to unlearn up to 50 classes.
This demonstrates our ACU maintains its high-efficiency advantage from the sample-level unlearning scenario in the class-level unlearning setting.

\subsection{Key Takeaways}

For unlearning effectiveness, our ACU achieves exact unlearning, consistently attaining optimal performance across all unlearning effectiveness metrics under various settings.
In contrast, other methods can only reach suboptimal performance and often face trade-offs, i.e., optimizing one metric at the expense of another.
As for model fidelity, existing baselines suffer from catastrophic degradation when continually processing multiple CU requests, with the extent of collapse worsening as the number of requests increases.
Although some baselines appear to preserve retained set accuracy, this is typically achieved through fine-tuning with access to retained data, which violates the constraint of not accessing historical data during the CU phase, constituting a form of \textit{cheating}.
In contrast, ACU enables exact unlearning without compromising any aspect of model fidelity, and crucially, without replaying any retained data.
As for unlearning efficiency, our ACU also demonstrates significant superiority, achieving a 50--125× speedup over most baselines and more than a 10,000× speedup compared to re-training from scratch.
Moreover, the efficiency of our ACU exhibits significantly greater robustness to increasing numbers of CU requests compared to the baselines.

In summary, our ACU achieves efficient and exact forgetting while preserving data privacy.
Its remarkable advantages in unlearning effectiveness, model fidelity, and system efficiency make it particularly well-suited for the CU setting.
The superiority of our ACU underscores the potential of gradient-free methods for unlearning and can inspire further promising exploration in this direction.

\section{Conclusion and Discussion}
\label{sec:conclusion}

In this paper, we present a practical problem, namely CU, which aims to sequentially forget the knowledge acquired during the CL phase.
We analyze the key challenges for CU and identify the fundamental issue of existing methods as their reliance on gradients.
Moreover, drawing inspiration from the emergence of gradient-free techniques (i.e., ACL methods) in the CL community, we propose our ACU to explore the gradient-free power in CU.
To the best of our knowledge, our ACU is the first work to fulfill efficient and exact forgetting with historical data privacy preservation.
Moreover, our ACU can be seen as a plug-and-play solution for unlearning in existing state-of-the-art ACL methods, without replaying any historical data.
By incorporating unlearning into ACL, our ACU holds great potential in boosting better trustworthiness and responsibility in CL.

Since the PTM backbone is typically frozen during the CL phase (as a common practice in existing state-of-the-art CL methods), our ACU primarily focuses on the unlearning for the analytic classifier.
Meanwhile, considering forgetting the knowledge within the PTM backbone is equally important, which is typically termed Continual Forgetting (CF)~\cite{Continual-forgetting,Continual-forgetting2}.
Yet, CF is actually orthogonal to our proposed CU.
Specifically, CF solely involves forgetting knowledge from the pre-training phase in a centralized manner, while our CU focuses on the knowledge acquired during the online CL phase under privacy constraints.
Considering the unlearning of both the PTM backbone and the analytic classifier would be an interesting and promising direction for future work.

\bibliographystyle{IEEEtran}
\bibliography{main}

\end{document}